\documentclass[10pt,twocolumn,letterpaper]{article}

\usepackage[pagenumbers]{cvpr} 
\usepackage{microtype}
\usepackage{graphicx}
\usepackage{booktabs} 
\usepackage{makecell}
\usepackage{multirow}
\usepackage{float}
\usepackage{algorithm}
\usepackage{algorithmic}
\newtheorem{thm}{Theorem}[section]
\usepackage[dvipsnames]{xcolor}

\usepackage{amsmath,amsfonts,bm}

\def\eqref#1{equation~\ref{#1}}

\def\1{\bm{1}}

\def\ry{{\textnormal{y}}}

\def\rvv{{\mathbf{v}}}

\def\rvx{{\mathbf{x}}}

\def\rvz{{\mathbf{z}}}

\def\vtheta{{\bm{\theta}}}

\DeclareMathAlphabet{\mathsfit}{\encodingdefault}{\sfdefault}{m}{sl}
\SetMathAlphabet{\mathsfit}{bold}{\encodingdefault}{\sfdefault}{bx}{n}

\newcommand{\E}{\mathbb{E}}

\newcommand{\R}{\mathbb{R}}

\DeclareMathOperator*{\argmax}{arg\,max}

\definecolor{cvprblue}{rgb}{0.21,0.49,0.74}
\usepackage[pagebackref,breaklinks,colorlinks,citecolor=cvprblue]{hyperref}

\title{Adversarially Robust Industrial Anomaly Detection Through Diffusion Model}

\author{Yuanpu Cao, Lu Lin, Jinghui Chen  \\
The Pennsylvania State University\\
\texttt{\{ymc5533,lulin,jzc5917\}@psu.edu} 
}

\begin{document}
\maketitle
\begin{abstract}
Deep learning-based industrial anomaly detection models have achieved remarkably high accuracy on commonly used benchmark datasets. However, the robustness of those models may not be satisfactory due to the existence of adversarial examples, which pose significant threats to the practical deployment of deep anomaly detectors. 
Recently, it has been shown that diffusion models can be used to purify the adversarial noises and thus build a robust classifier against adversarial attacks. Unfortunately,  we found that naively applying this strategy in anomaly detection (i.e., placing a purifier before an anomaly detector) will suffer from a high anomaly miss rate since the purifying process can easily remove both the anomaly signal and the adversarial perturbations, causing the later anomaly detector failed to detect anomalies. To tackle this issue, we explore the possibility of performing anomaly detection and adversarial purification simultaneously. We propose a simple yet effective adversarially robust anomaly detection method, \textit{AdvRAD}, that allows the diffusion model to act both as an anomaly detector and adversarial purifier. We also extend our proposed method for certified robustness to $l_2$ norm bounded perturbations. Through extensive experiments, we show that our proposed method exhibits outstanding (certified) adversarial robustness while also maintaining equally strong anomaly detection performance on par with the state-of-the-art methods on industrial anomaly detection benchmark datasets.
\end{abstract}    
\section{Introduction}
\label{sec:intro}
Anomaly detection aims at identifying data instances that are inconsistent with the majority of data, which has been widely applied in large-scale industrial manufacturing~\citep{bergmann2019mvtec} where efficient automatic anomaly detectors are deployed to spot diverse defects of industrial components varying from detecting scratches and leakages in capsules to finding impaired millimeter-sized components on a complicated circuit board ~\citep{zou2022spot}. Recently, deep learning (DL) based anomaly detection methods have achieved remarkable improvement over traditional anomaly detection strategies~\citep{ruff2021unifying,pang2021deep}. DL-based methods take advantage of neural networks to estimate the \textit{anomaly score} of a data instance which reflects how likely it is an anomaly. One common practice defines anomaly score as the \textit{reconstruction error} between the original data instance and the recovered one decoded by a symmetric neural network model (e.g., autoencoder)~\citep{hawkins2002outlier,chen2017outlier}. The insight that the reconstruction error can serve as the anomaly score is that the model trained on normal data usually cannot reproduce anomalous instances~\citep{bergmann2021mvtec}, thus a high reconstruction error for a data instance indicates a larger probability of it being an anomaly.

Though DL-based anomaly detection methods have achieved remarkably high accuracy on commonly used benchmark datasets~\citep{yu2021fastflow,lee2022cfa}, the robustness of the detection models is still unsatisfactory due to the existence of adversarial examples~\citep{goodge2020robustness, lo2022adversarially}, which poses significant threats to the practical deployment of deep anomaly detectors. An imperceptible perturbation on the input data could cause a well-trained anomaly detector to return incorrect detection results. Specifically, an anomalous instance, when added with an invisible noise, could cheat the detector to output a low anomaly score; while the normal instance can also be perturbed to make the detector raise a false alarm with a high anomaly score. In fact, such a robustness issue is not unique to a specific model, but a common problem for various state-of-the-art deep anomaly detection models (as will be seen in our later experiments in Section \ref{sec:attack}).

\begin{figure*}[tb]
\begin{center}
\centerline{\includegraphics[width=1.0\textwidth]{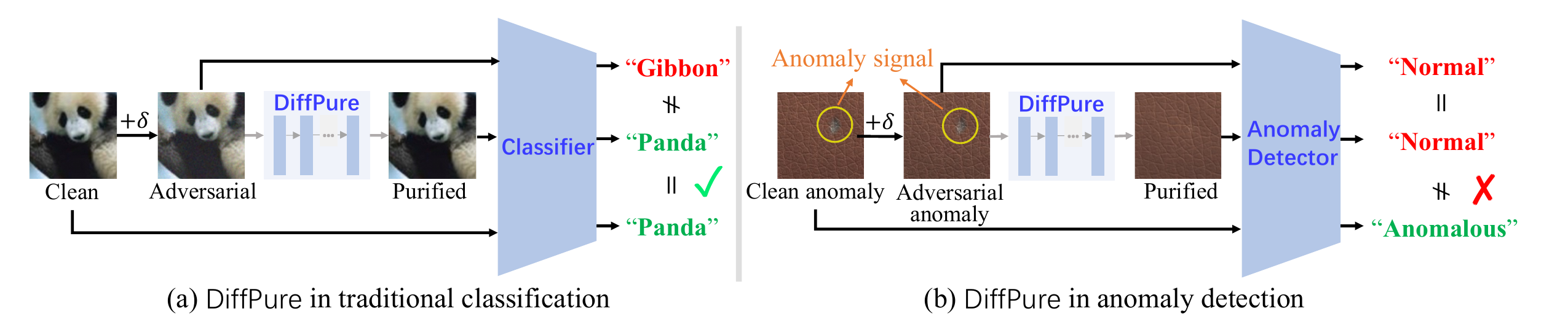}}
\vskip -0.1in
\caption{\textbf{Left}: purification-based adversarial robust model in the traditional classification task. \textbf{Right}: purification-based adversarial robust anomaly detection model. The anomaly signal can also be erased during the purification process leading to a high anomaly miss rate.}
\label{fig:comp_purify_ours}
\vskip -0.4in
\end{center}
\end{figure*}

Recently, \citet{nie2022DiffPure} have shown that diffusion models~\citep{ho2020denoising, song2021scorebased} can be used as data purifier to mitigate adversarial noises, and the proposed DiffPure \citep{nie2022DiffPure} achieves state-of-the-art defense performance. As a powerful class of generative models, diffusion models{~\citep{ho2020denoising,nichol2021improved}} are capable of generating samples with high quality, beating GANs in image synthesis~\citep{dhariwal2021diffusion}. Specifically, diffusion models first gradually add random noise and convert the data into standard Gaussian noise, and then learn the generative process to reverse the process and generate samples by denoising one step at a time. The denoising capability of diffusion models makes it possible to use it against imperceptible adversarial perturbations. As shown in Figure \ref{fig:comp_purify_ours} (left), DiffPure \citep{nie2022DiffPure} constructs a robust classifier by leveraging the diffusion model to purify adversarially perturbed images before classification. However, in anomaly detection scenario, naively placing DiffPure \citep{nie2022DiffPure} before another anomaly detector will largely deteriorate the detection performance as the purifier can also purify the anomaly signals along with the adversarial perturbations. Figure \ref{fig:comp_purify_ours} (right) shows a simple case of how DiffPure \citep{nie2022DiffPure} fails in anomaly detection. We can observe that when receiving an ``leather'' image with both imperceptible adversarial noise and anomaly signal (i.e., color defects), DiffPure \citep{nie2022DiffPure} essentially erase the color defects along with the adversarial perturbations, leading to high anomaly miss rate\footnote{We will discuss more experiments in Section \ref{sec:naive_attempt}.}.  

The key reason behind the failure of directly applying DiffPure \citep{nie2022DiffPure} in anomaly detection lies in that the purifying process can easily remove both the anomaly signal and the adversarial perturbations. While in the ideal case, DiffPure should only remove the adversarial perturbation while preserving the anomaly signal for accurate detection later. Given this observation, a natural question arises: 
\begin{center}
\textit{Is it possible to develop a method to simultaneously perform anomaly detection and adversarial purification together?} 
\end{center}
If the answer is yes, we don't need to enforce the purifier to distinguish between anomaly signals and adversarial perturbations, which is quite challenging. 
Based on this motivation, we explore the possibility of making the diffusion model act both as an \textit{anomly detector} and \textit{adversarial purifier} simultaneously and propose a novel adversarially robust anomaly detection method, termed \textit{AdvRAD}.

We summarize our contributions as follows:
\begin{itemize}
    \item We build a unified adversarial attack framework for various kinds of anomaly detectors to facilitate the adversarial robustness study in industrial anomaly detection domain, through which we systematically evaluate the adversarial robustness of state-of-the-art deep anomaly detection models. 
    \item We propose a novel adversarially robust industrial anomaly detection model through the diffusion model, inside which the diffusion model simultaneously performs anomaly detection and adversarial purification. We also extend our method for certified robustness to $l_2$ norm perturbations through randomized smoothing which provides additional robustness guarantees.
    \item We conduct extensive experiments and show that our method exhibits outstanding (certified) adversarial robustness, while also maintaining equally strong anomaly detection performance on par with the state-of-the-art anomaly detectors on industrial anomaly detection benchmark datasets MVTec AD~\citep{bergmann2019mvtec}, ViSA~\citep{zou2022spot}, and BTAD~\citep{mishra2021vt}.
\end{itemize}

\section{Related Work}
\noindent\textbf{Anomaly Detection Methods } Existing anomaly detection methods can be roughly categorized into two kinds: \emph{reconstruction-based} and \emph{feature-based}. One commonly used \emph{reconstruction-based} approach for anomaly detection is to train the autoencoder and use the $l_p$ norm distance between input and its reconstruction as the anomaly score~\citep{hawkins2002outlier,chen2017outlier,zhou2017anomaly}. \citet{bergmann2018improving} replace $l_p$ distance with SSIM~\citep{wang2004image} to have a better measure for perceptual similarity. Another more advanced branch of reconstruction-based models combines autoencoder with GAN, where the generator of the GAN is implemented using autoencoder~\citep{hou2021divide,liang2022omni,akccay2019skip}. These methods additionally incorporate the anomaly score with the similarity between the features of the input and the reconstructed images extracted from the discriminator to boost performance on categories that are difficult to reconstruct accurately. \emph{Feature-based} methods use pre-trained Resnet and vision transformer~\citep{yu2021fastflow}, or pre-trained neural networks with feature adaptation~\citep{lee2022cfa} to extract discriminative features for normal images, and estimate the distribution of these normal features by Flow-based model~\citep{gudovskiy2022cflow,rudolph2022fully}, KNN~\citep{reiss2021panda}, or Gaussian distribution modeling~\citep{li2021cutpaste}. These methods calculate the anomaly score using the distance from the features of test images to the established distribution for features of normal images.

\noindent\textbf{Adversarial Attacks and Defenses for Anomaly Detectors } To the best of our knowledge, existing attack and defense strategies for anomaly detectors only focus on autoencoder-based models. \citet{goodge2020robustness} consider perturbations to anomalous data that make the model categorize them as the normal class by reducing reconstruction error. For defense, they propose APAE using approximate projection and feature weighting to improve adversarial robustness. \citet{lo2022adversarially} extend the similar attack strategy to both normal and anomalous data and propose Principal Latent Space as a defense strategy to perform adversarially robust novelty detection. While they achieve a certain level of robustness, their performances on clean anomaly detection tasks are yet far from satisfactory.

\noindent\textbf{Diffusion Models } As a class of powerful generative models, diffusion models have attracted the most recent attention due to their high sample quality and strong mode coverage~\citep{sohl2015deep,ho2020denoising,nichol2021improved}. Recently, \citet{nie2022DiffPure} use diffusion models to purify adversarial perturbations for downstream robust classification, and the proposed DiffPure presents empirically strong robustness. In medical diagnostics, \citet{wolleb2022diffusion} adopt deterministic DDIM~\citep{song2020denoising} for supervised brain tumor detection. \citet{wyatt2022anoddpm} solve the same task under an unsupervised scenario using DDPM~\citep{ho2020denoising} with specially designed simplex noise for tumorous dataset. Note that these diffusion-based medical anomaly detection methods focusing on pixel-level anomaly are not directly comparable to our image-level anomaly detection. Moreover, none of the previous works have studied to improve the adversarial robustness of anomaly detection through diffusion models. We defer more comparison with the above diffusion model-based anomaly detector in Section \ref{sec:anoddpm}.
\section{Building Unified Adversarial Attacks for Anomaly Detectors}
\label{sec:attack}
To facilitate the adversarial robustness study on various kinds of anomaly detectors, we first build a unified adversarial attack framework in the context of anomaly detection. We consider the adversarial perturbations to be imperceptible, i.e., their existence will not flip the ground truth of the image (normal or anomalous).
The general goal of the unified attack is to make detectors return incorrect detection results by reducing anomaly scores for anomalous samples and increasing anomaly scores for normal samples. 
In particular, we use Projected Gradient Descent (PGD)~\citep{madry2018towards} to build the attack.

\noindent\textbf{PGD Attack on Anomaly Detector } Consider a sample $\rvx \in \R^d$ from the test dataset with label $\ry \in\{-1, 1\}$ (where ``$-1$'' denotes the anomalous class and ``$1$'' indicates the normal class), and a well-trained anomaly detector $A_{\vtheta}: \R^d \rightarrow \R$ that computes an anomaly score for each data sample. We define the optimization objective of PGD attack on the anomaly detector as:
    $\argmax_{\rvx} L_{\vtheta}(\rvx, \ry) = \ry A_{\vtheta}(\rvx),$
where $\ry$ guides the direction of perturbing $\rvx$ to increase or decrease its anomaly score. Depending on the perturbation constraint, adversarial examples can be generated by $l_{\infty}$-norm or $l_{2}$-norm bounded PGD, respectively as:
\begin{align}
	&\rvx_{n+1} = P_{\rvx, \epsilon}^{l_{\infty}}\left\{\rvx_{n} + \alpha \cdot\text{sgn}(\nabla_{\rvx_{n}} L_{\vtheta}(\rvx_n, \ry)\right\} \\
 	&\rvx_{n+1} = P_{\rvx, \epsilon}^{l_{2}}\left\{\rvx_{n} + \alpha \frac{\nabla_{\rvx_{n}}L_{\vtheta}(\rvx_n, \ry)}{\left\| \nabla_{\rvx_{n}}L_{\vtheta}(\rvx_n, \ry) \right\|} \right\}
\end{align}
where $\alpha$ is the step size, $n\in [0, N-1]$ is the current step of in total $N$ iterations, and $\rvx_{0}=\rvx$. $P_{\rvx, \epsilon}^{l_{p}}\{\cdot\}$ denotes the projection on $\rvx_{n+1}$ such that $\left\| \rvx_{n+1} - \rvx \right\|_{p} \leq \epsilon$.
The final adversarial example is generated by $\rvx_{adv}=\rvx_{N}$. This attacking strategy encapsulates previous works on adversarial examples for anomaly detectors, where only autoencoder-based models were considered~\citep{lo2022adversarially,goodge2020robustness}. The anomaly score can be specified as $\displaystyle A_{\vtheta}(\rvx) = \left\| D(E(\rvx)) - \rvx \right\|$ to accommodate to their scenarios, where $D$ denotes the decoder and $E$ corresponds to the encoder. 

\noindent\textbf{Robustness Evaluation on Existing Anomaly Detectors }
Based on the unified PGD attack, we systematically evaluate the adversarial robustness of the state-of-the-art detectors with various models.
Table \ref{table:exp_demo2} demonstrates the efficacy of the attack in disclosing the vulnerability of existing anomaly detectors: the AUC scores of these advanced anomaly detectors drop to as low as $0$ under adversarial perturbations with $l_\infty$ norm less than $2/255$ on \emph{Toothbrush} dataset from benchmark MVTec~\citep{bergmann2019mvtec}. This suggests that current anomaly detectors suffer from fragile robustness on adversarial data.

\begin{table}[ht]
\begin{center}
\resizebox{1.0\linewidth}{!}{
    \begin{tabular}{l c c c c c}
    \toprule
        Method & OCR-GAN \citep{liang2022omni} & SPADE \citep{cohen2020sub} & CFlow \citep{gudovskiy2022cflow} & FastFlow \citep{yu2021fastflow} & CFA \citep{lee2022cfa} \\
    \midrule
    Standard AUC & $96.7$ & $88.9$ & $85.3$ & $94.7$ & $100$ \\
    Robust AUC & $0$ & $0$ & $0$ & $0$ & $0$ \\
    \bottomrule
    \end{tabular}}
\end{center}
\vskip -0.15in
\caption{Standard AUC and robust AUC against $l_{\infty}$-PGD ($\epsilon=2/255$) attacks on \emph{Toothbrush} dataset from benchmark MVTec AD, obtained by various anomaly detection SOTAs.}
\label{table:exp_demo2}
\vskip -0.2in
\end{table}
\section{Adversarially Robust Anomaly Detection} \label{sec:defense}
Before we introduce our novel robust anomaly detection method, we first give a brief 
review of diffusion models \citep{sohl2015deep,ho2020denoising,nichol2021improved} and present a naive attempt of applying DiffPure \citep{nie2022DiffPure} on anomaly detection and analyze its failure case.
\begin{table*}[ht]
\begin{center}
\small
\resizebox{0.8\linewidth}{!}{
    \begin{tabular}{c  c c c c c c}
    \toprule
        Purification-level & $p=5$  & $p=25$ & $p=50$ & $p=100$ & $p=200$ & $p=300$ \\
        \midrule
        Standard AUC & $93.6$ & $95.0$ & $87.5$ & $74.7$ & $47.5$ & $36.4$ \\
        Robust AUC & $32.2 (\downarrow 61.4)$ & $47.8 (\downarrow 47.2)$ & $50.6 (\downarrow 36.9) $ & $34.2 (\downarrow 40.5)$ & $9.4 (\downarrow 38.1)$ & $19.7 (\downarrow 16.7)$ \\
    \bottomrule
    \end{tabular}}
\end{center}
\vskip -0.15in
\caption{Standard AUC and robust AUC against $l_{\infty}$-PGD $(\epsilon=2/255)$ attacks on \emph{Toothbrush} dataset, obtained by DiffPure \citep{nie2022DiffPure} + CFA \citep{lee2022cfa} with varying purification-level $p$ (diffusion steps, max=1000).}
\label{table:exp_purify_demo}
\vskip -0.2in
\end{table*}
\subsection{Preliminaries on Diffusion Models}  \label{sec:prelim}
We follow the formulation of DDPMs given in \citep{ho2020denoising, nichol2021improved}, which defines a $T$ steps diffusion process $q(\rvx_{1:T}|\rvx_0):=\prod_{t=1}^T q(\rvx_t|\rvx_{t-1})$ parameterized by a variance schedule $\beta_1,\ldots,\beta_T$ as $q(\rvx_t|\rvx_{t-1}):=\mathcal{N}(\rvx_t;\rvx_{t-1}\sqrt{1-\beta_t}, \beta_t I)$,
which iteratively transforms an unknown data distribution ${q(\rvx_0)}$ to standard Gaussian ${q(\rvx_T)}=\mathcal{N}(0, \mathbf{I})$.
The generative process $p_\vtheta(\rvx_{0:T}) := p(\rvx_T)\prod_{t=1}^Tp_\vtheta(\rvx_{t-1}|\rvx_t)$ is learned to approximate each $q(\rvx_{t-1}|\rvx_t)$ using neural networks as $p_\vtheta(\rvx_{t-1}|\rvx_t):= \mathcal{N}(\rvx_{t-1};\boldsymbol{\mu}_\vtheta(\rvx_t,t), \boldsymbol{\Sigma}_\vtheta(\rvx_t, t))$
A noticeable property of the diffusion process is that it allows directly sampling $\rvx_t$ at an arbitrary timestep $t$ given $\rvx_0$. Using the notation $\alpha_t:=1-\beta_t$ and $\overline{\alpha_t}:=\prod_{s=1}^t\alpha_s$, we have
\begin{equation}
    \rvx_t = \sqrt{\overline{\alpha}_t}\rvx_0 + \sqrt{1-\overline{\alpha}_t}\bm{\epsilon}, \quad \bm{\epsilon} \in \mathcal{N}(0, \mathbf{I}) \label{eq:diffusion}
\end{equation}
For training the diffusion model, \citet{ho2020denoising} propose a simplified objective without learning signals for $\boldsymbol{\Sigma}_\vtheta(\rvx_t,t)$: $L_{\text{simple}} = \E_{t, \rvx_0, \bm{\epsilon}}[\| \bm{\epsilon} - \bm{\epsilon}_{\vtheta}(\rvx_{t}, t) \|]$
In this paper, we follow \citet{nichol2021improved} and train the diffusion model using a hybrid loss for better sample quality with fewer generation steps. More details can be found in Appendix \ref{sec:loss}.

\subsection{Naive Attempt: Applying DiffPure on Anomaly Detection} \label{sec:naive_attempt}
DiffPure \citep{nie2022DiffPure} uses the diffusion model to purify adversarially perturbed images before classification and present strong empirical robustness. A naive idea would be applying DiffPure in anomaly detection for better robustness. However, as mentioned previously, naively placing a purifier before another anomaly detector will largely deteriorate the detection performance as the purifier can also purify the anomaly signals along with the adversarial perturbations. For this strategy to work, DiffPure should only remove the adversarial perturbation while preserving the anomaly signal for anomaly detection later. Unfortunately, this is extremely difficult to achieve. 
To verify this, we directly apply DiffPure \citep{nie2022DiffPure} upon CFA \citep{lee2022cfa}, one of the SOTA anomaly detectors, with different purification levels (i.e., diffusion steps in DiffPure \citep{nie2022DiffPure}) and present the results in Table \ref{table:exp_purify_demo}. We can observe that, with a lower purification level (e.g., 5, 25 diffusion steps), this method maintains high standard AUC while the robust AUC is far from satisfactory, suggesting that it is unable to fully remove the adversarial perturbations; when increasing the purification level, standard AUC will rapidly decrease, suggesting that the anomaly signal was also removed. 
This observation motivates us to build adversarially robust anomaly detectors that can simultaneously perform anomaly detection and adversarial purification.

\subsection {Merging Anomaly Detection and Adversarial Purification through Diffusion Model}
Observing the failure case of naively applying DiffPure on anomaly detection, it is natural to ask whether we can simultaneously perform anomaly detection and 
adversarial purification together. If so, we can avoid enforcing the purifier to distinguish between anomaly signals and adversarial perturbations. Surprisingly, we found a simple yet effective strategy to merge anomaly detection and adversarial purification tasks into one single \textit{robust reconstruction} procedure through the diffusion model. We named this adversarially robust anomaly detection method \textit{AdvRAD}. 

\begin{figure}[ht]
\begin{center}
\centerline{\includegraphics[width=0.85\columnwidth]{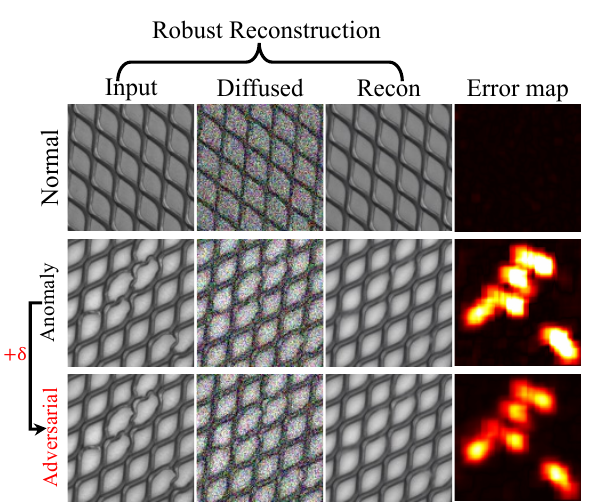}}
\vskip -0.1in
\caption{Reconstruction results of normal data, anomalous data, and adversarially perturbed data using our model. The observed reconstruction is robust to adversarial noise.}
\label{fig:case}
\end{center}
\vskip -0.3in
\end{figure}

\noindent\textbf{Robust Reconstruction }
Robust reconstruction in \textit{AdvRAD} is the key to merging anomaly detection and adversarial purification tasks together. It relies on the fact that the diffusion model itself can be used as a reconstruction model and the reconstruction error can be used as a natural anomaly score.  Specifically, the diffusion model training procedure is essentially predicting noise added in the diffusion process and then denoising. Instead of using the trained diffusion model to generate new samples through multiple denoising steps from a noise, one can also start from an original image, gradually add noise and then denoise to reconstruct the original image. Since the diffusion model is trained on the normal data samples, such reconstruction error can serve as a natural indicator of the anomaly score. 
In Figure \ref{fig:case}, we show an example of robust reconstruction using diffusion models. As can be seen from Figure \ref{fig:case}, for normal data, the reconstruction is nearly identical to the input. For anomaly data, the diffusion model (after adding noise and denoising) could ``repair'' the anomaly regions, thus obtaining high reconstruction error, which could be easily detected as anomalies. Now let's consider adversarial robustness in anomaly detection. 
From Section \ref{sec:naive_attempt}, we already know that the adversarial noise will be removed together with the anomaly signal by the diffusion model. Therefore, after robust reconstruction, the adversarially perturbed anomaly sample could still be recovered to normal case and thus obtain a high reconstruction error as shown in Figure \ref{fig:case}. In this way, \textit{AdvRAD} no longer needs to distinguish between adversarial noise and anomaly signal and thus only needs to remove both simultaneously. We summarize the robust construction steps in Algorithm \ref{alg:full-shot} in Appendix \ref{sec:full-shot}.  

To obtain the best performances of \textit{AdvRAD}, there are still several things to notice: 1) The diffusion steps $k$ in robust reconstruction should be chosen such that the amount of Gaussian noise is dominating the adversarial perturbations and anomaly signals while the high-level features of the input data are still preserved for reconstruction. 2) One major problem with the traditional diffusion denoising algorithm (see Algorithm \ref{alg:full-shot} in Appendix \ref{sec:full-shot}) is that the iterative denoising procedure is time-consuming, making it unacceptable for real-time anomaly detection in critical situations~\citep{sun2021ctf}. Moreover, extra reconstruction error can also be introduced due to the multiple sampling steps. To overcome these challenges, we investigate  the arbitrary-shot denoising process allowing fewer denoising steps, with the details shown in Appendix \ref{sec:arbitrary-shot}. Based on our results ({see Appendix \ref{sec:fewer_shot}}) we observe that one-shot denoising (Algorithm \ref{alg:one-shot}) is sufficient to produce an accurate reconstruction result with $\mathcal{O}(1)$ inference-time efficiency. Such a one-shot idea has also been adopted in \citet{carlini2022certified} for robust image classification. By default, we use one-shot robust reconstruction for all experiments in Section \ref{sec:exp}.
\begin{algorithm}[tb]
   \caption{One-shot Robust Reconstruction in \textit{AdvRAD}}
   \label{alg:one-shot}
\begin{algorithmic}[1]
   \STATE {\bfseries Input:} Test images: $\rvx$, diffusion steps: $k(k\leq T)$
   \STATE {\bfseries Output:} Reconstructions of $\rvx$: $\tilde{\rvx}$
   \STATE $\bm{\epsilon} \sim \mathcal{N}(0, \mathbf{I})$
   \STATE $\rvx_k = \sqrt{\overline{\alpha}_k}\rvx_0 + \sqrt{1-\overline{\alpha}_k}\bm{\epsilon}$ \\ // one-shot denoising process:
   \STATE $\tilde{\rvx} = \frac{1}{\sqrt{\overline{\alpha}_{k}}}({\rvx}_{k}- \sqrt{1-\overline{\alpha}_{k}}\bm{\epsilon}_{\vtheta}(\rvx_{k}, k))$
\end{algorithmic}
\end{algorithm}

\noindent\textbf{Anomaly Score Calculation:}
To calculate the final anomaly score in a robust and stable manner, we first calculate the Multiscale Reconstruction Error Map (denoted as $\text{Err}_{\text{ms}}$), which considers both pixel-wise and patch-wise reconstruction errors. Specifically, for each scale $l$ in $L=\{1, \frac{1}{2}, \frac{1}{4}, \frac{1}{8}\}$, we first calculate the error map ${\text{Err}(\rvx, \tilde{\rvx})}_l$ between the downsampled input ${\rvx}^l$ and the downsampled reconstruction ${\tilde{\rvx}}^l$ with  $\frac{1}{C}\sum_{c=1}^{C}{({\rvx}^l-{\tilde{\rvx}}^l)}_{[c,:,:]}^2$ where the square operator is abused here for element-wise square operation, then unsampled to the original resolution. The final $\text{Err}_{\text{ms}}$ is obtained by averaging each scale's error map and applying a mean filter for better stability similar to \citet{zavrtanik2021reconstruction}: 
    ${\text{Err}_{\text{ms}}(\rvx, \tilde{\rvx})} = (\frac{1}{N_L}\sum_{l \in L}{\text{Err}(\rvx, \tilde{\rvx})}_l) \ast f_{s \times s}$
where $f_{s \times s}$ is the mean filter of size ${s \times s}$, $\ast$ is the convolution operation. Similar to \citet{pirnay2022inpainting}, we take the pixel-wise maximum of the absolute deviation of the $\text{Err}_{\text{ms}}(\rvx, \tilde{\rvx})$ on normal training data as the scalar anomaly score. Due to space limits, we leave the complete anomaly score calculation algorithm in Appendix \ref{sec:score}.
\section{Experiments}\label{sec:exp}
We compare our proposed \textit{AdvRAD} with state-of-the-art anomaly detectors on both clean input and adversarially perturbed input. \textit{AdvRAD} shows a stronger robustness performance compared with SOTAs even combined with model-agnostic defenses (i.e., DiffPure \citep{nie2022DiffPure} and Adversarial Training \citep{madry2018towards}) and domain-specific defense-enabled anomaly detector baselines, and also maintains robustness even under stronger adaptive attacks. Finally, we further extend \textit{AdvRAD} for certified robustness to $l_2$ norm perturbations.

\subsection{Experimental Settings} \label{sec:exp_settings}
\textbf{Dataset and Model Implementation:} We perform experiments on three industrial anomaly detection benchmark datasets MVTec AD ~\citep{bergmann2019mvtec}, ViSA ~\citep{zou2022spot} and BTAD ~\citep{mishra2021vt} datasets. MVTec AD comprises 15 different texture (e.g., leather, wood) and object (e.g., toothbrush, transistor) categories which showcase more than $70$ types of anomalies from the real world. ViSA covers 12 objects with challenging scenarios including complex structures, multiple instances and object pose/location variations. BTAD contains $3$ industrial products showcasing body and surface anomalies.  We resize all images to 256$\times$256 resolution in our experiments. We implement the diffusion model based on \citet{nichol2021improved} using U-Net backbone~\citep{ronneberger2015u}. We set the total iteration step as $T=1000$ for all experiments. During the inference stage, we choose the diffusion step $k = 100$ in our experiments (see Appendix \ref{sec:k} for sensitivity test). More hyperparameters are described in Appendix \ref{sec:hyper}.

\noindent\textbf{Adversarial Attacks:} 
We adopt commonly used PGD attack~\citep{madry2018towards} to compare with the state-of-the-art anomaly detection models and defense-enabled anomaly detectors. Additionally, we also consider the BPDA, EOT attack~\citep{athalye2018obfuscated}, and AutoAttack \citep{croce2020reliable} for better robustness evaluations on defense-enabled anomaly detectors and ours. We set the attack strength $\epsilon=2/255$ for $l_\infty$-norm  attacks and $\epsilon=0.2$ for $l_2$-norm attacks to ensure imperceptible attack perturbations. 

\noindent\textbf{Evaluation Metric:} We use the widely-adopted AUC (area under the receiver operating characteristic curve) to evaluate anomaly detection performance. Specifically, we consider \emph{standard AUC} and \emph{robust AUC}. The standard AUC evaluates the performance on the clean test data, while the robust AUC evaluates the performance on adversarially perturbed data.

\subsection{Comparison with the State-of-the-art Anomaly Detectors} 
\label{sec:comp_sota}
We compare our method \textit{AdvRAD} with five state-of-the-art methods for image anomaly detection: SPADE~\citep{cohen2020sub}, OCR-GAN~\citep{liang2022omni}, CFlow~\citep{gudovskiy2022cflow}, FastFlow~\citep{yu2021fastflow}, and CFA~\citep{lee2022cfa}, against the $l_{\infty}$-PGD and $l_{2}$-PGD attacks. Table \ref{table:exp_sota_linf}, \ref{table:exp_sota_linf_visa} and \ref{table:exp_sota_linf_btad} present the robustness performance against $l_{\infty}$-PGD attacks ($\epsilon=2/255$) on MVTec AD, ViSA and BTAD, respectively. We observe that our method largely outperforms previous methods regarding robust AUC against $l_{\infty}$-PGD attacks ($\epsilon=2/255$). Specifically, our method improves robust AUC on all $15$ categories of MVTec AD and obtains the average robust AUC of $81.1\%$ with the improvement of at least $78.8\%$. In ViSA and BTAD, our method improves the avg robust AUC by $80.6\%$ and $72.6\%$, respectively. See Appendix \ref{sec:l2} for similar results against $l_{2}$-PGD attacks ($\epsilon=0.2$). In the meantime, we can observe that in terms of anomaly detection performance on clean data, the average standard AUC obtained by our method is on par with the state-of-the-art methods in MVTec AD and BTAD datasets, while beating all baselines in ViSA. These results clearly demonstrate the effectiveness of our proposed method in defending against $l_{\infty}$-PGD and $l_{2}$-PGD attacks, while also maintaining strong anomaly detection performance on benchmark datasets.

\begin{table}[t]
\begin{center}
\resizebox{1.0\linewidth}{!}{
    \begin{tabular}{ c| c c c c c | c c }
    \toprule
       Category & OCR-GAN  & SPADE & CFlow & FastFlow  & CFA & AdvRAD \\
      \midrule
       Carpet & $0^{(76.6)}$ & $0^{(92.8)}$ & $0^{(98.6)}$ & $0^{(\mathbf{99.7})}$ & $0^{(99.4)}$ & $\mathbf{70.5}^{(73.8)}$ \\
      Grid & $0^{(97)}$ & $0^{(47.3)}$ & $0^{(96.6)}$ & $0^{(\mathbf{100})}$ & $0^{(99.6)}$ & $\mathbf{99.6}^{(\mathbf{100})}$ \\ 
      Leather & $0^{(90.7)}$ & $0^{(95.4)}$ & $0^{(\mathbf{100})}$ & $6.6^{(\mathbf{100})}$ & $2.0^{(\mathbf{100})}$ & $\mathbf{97.8}^{(\mathbf{100})}$ \\
      Tile & $0^{(95.6)}$ & $0^{(96.5)}$ & $0^{(99.6)}$ & $1.3^{(\mathbf{100})}$ & $0.1^{(99.3)}$ & $\mathbf{93.9}^{(95.4)}$ \\
      Wood & $0^{(95.4)}$ & $0^{(95.8)}$ & $0^{(99.7)}$ & $0^{(\mathbf{99.9})}$ & $0^{(99.7)}$ & $\mathbf{95.2}^{(98.3)}$ \\\midrule
       Bottle & $0^{(97.7)}$ & $0^{(97.2)}$ & $0^{(\mathbf{100})}$ & $0^{(\mathbf{100})}$ & $0.1^{(\mathbf{100})}$ & $\mathbf{96.9}^{(99.6)}$ \\
      Cable & $0^{(71.5)}$ & $0^{84.8)}$ & $0^{(98.7)}$ & $0^{(67.4)}$ & $0.8^{(\mathbf{99.8})}$ & $\mathbf{38.9}^{(79.2)}$ \\ 
      Capsule & $0^{(80.4)}$ & $0^{(89.7)}$ & $0^{(93.7)}$ & $8.9^{(\mathbf{99.2})}$ & $0^{(97)}$ & $\mathbf{53.5}^{(90.5)}$ \\ 
      Hazelnut & $0^{(97.7)}$ & $0^{(88.1)}$ & $0^{(99.9)}$ & $0^{(99.5)}$ & $0.1^{(\mathbf{100})}$ & $\mathbf{91.5}^{(97.3)}$ \\
      Metal Nut & $0^{(82.6)}$ & $0^{(71)}$ & $0^{(\mathbf{100})}$ & $0^{(98.2)}$ & $0^{(\mathbf{100})}$ & $\mathbf{85.9}^{(91.0)}$ \\
      Pill & $0^{(80.8)}$ & $0^{(80.1)}$ & $0^{(93.2)}$ & $0^{(97.8)}$ & $0^{(\mathbf{98})}$ & $\mathbf{39.0}^{(94.4)}$ \\ 
      Screw & $0^{(\mathbf{99.4})}$ & $0^{(66.7)}$ & $0^{(79)}$ & $6.6^{(91.1)}$ & $0^{(95.5)}$ & $\mathbf{87.6}^{(99.3)}$ \\ 
      Toothbrush & $0^{(96.7)}$ & $0^{(88.9)}$ & $0^{(85.3)}$ & $0^{(94.7)}$ & $0^{(\mathbf{100})}$ & $\mathbf{95.8}^{(\mathbf{100})}$ \\
      Transistor & $0^{(75)}$ & $0^{(90.3)}$ & $0^{(98.3)}$ & $0^{(99.4)}$ & $0^{(\mathbf{100)}}$ & $\mathbf{74.5}^{(90.6)}$ \\
      Zipper & $0^{(80.4)}$ & $0^{(96.6)}$ & $0^{(97.5)}$ & $17.5^{(99.6)}$ & $0^{(99.7)}$ & $\mathbf{96.2}^{(\mathbf{99.7})}$ \\
      \midrule
      \textbf{Average} & $0^{(87.8)}$ & $0^{(85.4)}$ & $0^{(96.0)}$ & $2.3^{(98.5)}$ & $0.2^{(\mathbf{99.2})}$ & $\mathbf{81.1}^{(93.9)}$ \\
    \bottomrule
    \end{tabular}}
\end{center}
\vskip -0.2in
\caption{Robust AUC$^{\text{(Standard AUC)}}$ against $l_{\infty}$-PGD attacks on MVTec AD, obtained by SOTAs and ours.}
\vskip -0.1in
\label{table:exp_sota_linf}
\end{table}

\begin{table}[t]
\begin{center}
\resizebox{0.9\linewidth}{!}{
    \begin{tabular}{ c | c c c c |  c }
    \toprule
      Category & SPADE & CFlow & FastFlow & CFA & AdvRAD \\
      \midrule
       PCB1  & $0^{(88.9)}$ & $0^{(95.0)}$ & $0^{(88.8)}$ & $1.6^{(92.7)}$ & $\mathbf{87.6}^{(\mathbf{95.9})}$ \\
       PCB2  & $0^{91.8)}$ & $0^{(87.3)}$ & $0.1^{(88.7)}$ & $1.5^{(93.7)}$ & $\mathbf{94.6}^{(\mathbf{99.4})}$ \\ 
       PCB3 & $0^{(81.1)}$ & $0^{(82.1)}$ & $0^{(87.3)}$ & $2.4^{(94.3)}$ & $\mathbf{96.9}^{(\mathbf{99.4})}$ \\
       PCB4 & $0^{(\mathbf{95.7})}$ & $0^{(97.8)}$ & $0^{(96.5)}$ & $0^{(\mathbf{99.6})}$ & $\mathbf{97.6}^{(99.3)}$ \\
       Capsules  & $0^{(63.9)}$ & $0^{(58.3)}$ & $0^{(74.3)}$ & $0^{(64.7)}$ & $\mathbf{73.1}^{(\mathbf{88.4})}$ \\
       Candle  & $0^{(89.0)}$ & $0^{(95.6)}$ & $0^{(\mathbf{95.8})}$ & $0^{(91.6)}$ & $\mathbf{63.3}^{(92.2)}$ \\ 
       Macaroni1 & $0^{(80.7)}$ & $0^{(84.3)}$ & $0^{(89.1)}$ & $0^{(87.5)}$ & $\mathbf{83.5}^{(\mathbf{99.0})}$ \\
       Macaroni2 & $0^{(60.5)}$ & $0^{(72.7)}$ & $0^{(72.4)}$ & $0^{(76.5)}$ & $\mathbf{70.6}^{(\mathbf{93.0})}$ \\
       Cashew  & $0^{(93.2)}$ & $0^{(93.9)}$ & $0^{(91.4)}$ & $0.1^{(\mathbf{94.0})}$ & $\mathbf{72.4}^{(\mathbf{97.3})}$ \\
       Chewing gum  & $0^{(95.3)}$ & $0^{(98.0)}$ & $0.4^{(\mathbf{99.1})}$ & $2.2^{(\mathbf{99.4})}$ & $\mathbf{80.1}^{(\mathbf{99.4})}$ \\ 
       Fryum & $0^{(91.3)}$ & $0^{(86.2)}$ & $2.0^{(\mathbf{93.8})}$ & $0.6^{(92.6)}$ & $\mathbf{77.2}^{(93.7)}$ \\
       Pipe fryum & $0^{(74.0)}$ & $0^{(97.6)}$ & $0^{(96.5)}$ & $0^{(96.1)}$ & $\mathbf{79.1}^{(\mathbf{98.3})}$ \\
      \midrule
      \textbf{Average} & $0^{(83.8)}$ & $0^{(87.4)}$ & $0.2^{(89.5)}$ & $0.7^{(90.2)}$  & $\mathbf{81.3}^{(\mathbf{96.3})}$ \\
    \bottomrule
    \end{tabular}}
\end{center}
\vskip -0.2in
\caption{Robust AUC$^{\text{(Standard AUC)}}$ against $l_{\infty}$-PGD attacks on ViSA, obtained by SOTAs and ours.}
\vskip -0.2in
\label{table:exp_sota_linf_visa}
\end{table}

\begin{table}[t]
\begin{center}
\resizebox{0.85\linewidth}{!}{
    \begin{tabular}{ c | c c c c |  c }
    \toprule
      Category & SPADE & CFlow & FastFlow & CFA & AdvRAD \\
      \midrule
       01  & $0^{(94.5)}$ & $2.0^{(96.9)}$ & $5.6^{(99.4)}$ & $43.2^{(\mathbf{100})}$ & $\mathbf{100}^{(\mathbf{100})}$ \\
      02  & $0^{(77.0)}$ & $0^{(84.2)}$ & $22.4^{(\mathbf{84.5})}$ & $2.2^{(83.2)}$ & $\mathbf{63.9}^{(79.9)}$ \\ 
      03 & $0^{(\mathbf{100})}$ & $1.2^{(99.9)}$ & $7.0^{(94.5)}$ & $0.1^{(99.6)}$ & $\mathbf{99.5}^{(\mathbf{100})}$ \\
      \midrule
      \textbf{Average} & $0^{(90.5)}$ & $1.1^{(93.7)}$ & $11.7^{(92.8)}$ & $15.2^{(\mathbf{94.3})}$  & $\mathbf{87.8}^{(93.2)}$ \\
    \bottomrule
    \end{tabular}}
\end{center}
\vskip -0.2in
\caption{Robust AUC$^{\text{(Standard AUC)}}$ against $l_{\infty}$-PGD attacks on BTAD dataset, obtained by SOTAs and ours.}
\vskip -0.2in
\label{table:exp_sota_linf_btad}
\end{table}

\subsection{Comparisons with other Diffusion model based Anomaly Detectors} \label{sec:anoddpm}
 Current diffusion model-based anomaly detection methods primarily focus on medical images \citep{wyatt2022anoddpm, wolleb2022diffusion}. \citet{wolleb2022diffusion} adopt deterministic DDIM~\citep{song2020denoising} for supervised pixel-level tumor localization, where the ground truth of the image is required, which is fundamentally different from our unsupervised image-level anomaly detection. \citet{wyatt2022anoddpm} proposes AnoDDPM to solve the same task under an unsupervised scenario using DDPM \citep{ho2020denoising} with simplex noise. Note that AnoDDPM \citep{wyatt2022anoddpm} is also a pixel-level anomaly localization method which is not directly comparable to our image-level anomaly detection. To make the comparison with AnoDDPM \citep{wyatt2022anoddpm}, we use two image-level anomaly score calculation methods to convert it to an image-level anomaly detector: (1) the maximum pixel anomaly score (i.e., the maximum square error between the reconstruction and the initial image), and (2) our proposed anomaly score calculation as presented in Appendix \ref{sec:score}. 
 
 We summarize the (standard AUC, robust AUC) of AnoDDPM \citep{wyatt2022anoddpm} and our proposed \textit{AdvRAD} against $l_\infty$-PGD attacks on MVTec AD \citep{bergmann2019mvtec} in Table \ref{table:exp_anoddpm_linf} where AnoDDPM$^*$ refers to AnoDDPM with our proposed anomaly score calculation. We can clearly observe that our proposed \textit{AdvRAD} outperforms both AnoDDPM \citep{wyatt2022anoddpm} and AnoDDPM with the one-shot reconstruction in terms of standard AUC on clean data and robust AUC on perturbed data, which demonstrates that simplex noise cannot generalize well to industrial anomaly detection and deteriorate the adversarial robustness of the diffusion model in anomaly detection. Note that we only evaluate the robustness of AnoDDPM with our one-shot reconstruction for a fair comparison since \textit{AdvRAD} performs one-shot reconstruction. Moreover, we found that the one-shot technique also improves the performance of AnoDDPM \citep{wyatt2022anoddpm} on clean data. Although \citet{wolleb2022diffusion, wyatt2022anoddpm} have proposed using diffusion models for anomaly detection, and AnoDDPM \citep{wyatt2022anoddpm} adopt a similar algorithm framework with our proposed method, none of them have provided the understanding of diffusion model in anomaly detection from a robustness perspective. In addition, the experimental results evidently show that our proposed method outperforms AnoDDPM in terms of accuracy, efficiency, and adversarial robustness in industrial anomaly detection.

\begin{table}[t]
\begin{center}
\small
\resizebox{0.9\linewidth}{!}{
    \begin{tabular}{ c| c c c| c }
    \toprule
       Category & AnoDDPM & \makecell[c]{AnoDDPM \\ (one-shot)} & \makecell[c]{AnoDDPM$^*$ \\ (one-shot)} & AdvRAD \\
      \midrule
       Carpet & $(72.4, -)$ & ${(73.7, 20.2)}$ & ${(69.9, 9.7)}$ & $(\mathbf{73.8}, \mathbf{70.5})$ \\
      Grid & $(99.2, -)$ & ${(99.9, 81.6)}$ & ${(\mathbf{100}, 41.4)}$ & ${(\mathbf{100}, \mathbf{99.6})}$ \\ 
      Leather & $(81.8, -)$ & ${(91.9, 56.4)}$ & ${(93.4, 32.5)}$ &${(\mathbf{100}, \mathbf{97.8})}$ \\
      Tile & $(85.8, -)$ & ${(80.6, 38.7)}$ & ${(92.0, 16.2)}$ &${(\mathbf{95.4}, \mathbf{93.9})}$ \\
      Wood & $(90.2, -)$ & ${(94.8, 42.1)}$ & ${(94.1, 12.6)}$ &${(\mathbf{98.3}, \mathbf{95.2})}$ \\\midrule
      Bottle & $(94.8, -)$ & ${(95.6, 81.2)}$ & ${(97.1, 77.8)}$ &${(\mathbf{99.6}, \mathbf{96.9})}$ \\
      Cable & $(57.3, -)$ & ${(80.5, 1.4)}$ & ${(\mathbf{82.4}, 0.1)}$ &${(79.2, \mathbf{38.9})}$ \\ 
      Capsule & $(74.3, -)$ & ${(74.5, 47.8)}$ & ${(67.3, 33.9)}$ &${(\mathbf{90.5}, \mathbf{53.5})}$ \\ 
      Hazelnut & $(83.3, -)$ & ${(85.5, 13.9)}$ & ${(74.5, 4.4)}$ &${(\mathbf{97.3}, \mathbf{91.5})}$ \\
      Metal Nut & $(75.6, -)$ & ${(85.2, 9.6)}$ & ${(88.5, 3.7)}$ &${(\mathbf{91.0}, \mathbf{85.9})}$ \\
      Pill & $(59.4, -)$ & ${(60.7, \mathbf{49.8})}$ & ${(71.0, 31.7)}$ &${(\mathbf{94.4}, 39.0)}$ \\ 
      Screw & $(74.7, -)$ & ${(69.1, 0.3)}$ & ${(59.6, 0.9)}$ &${(\mathbf{99.3}, \mathbf{87.6})}$ \\ 
      Toothbrush & $(99.7, -)$ & ${(99.4, 14.7)}$ & ${(\mathbf{100}, 16.9)}$ &${(\mathbf{100}, \mathbf{95.8})}$ \\
      Transistor & $(68.5, -)$ & ${(64.6, 0.5)} $ & ${(70.9, 0.0)}$ & ${(\mathbf{90.6}, \mathbf{74.5})}$ \\
      Zipper & $(89.0, -)$ & ${(91.9, 72.2)}$ & ${(98.9, 64.1)}$ & ${(\mathbf{99.7}, \mathbf{96.2})}$ \\
      \midrule
      \textbf{Average} & $(80.4, -)$ & ${(83.2, 35.33)}$ & ${(84.0, 23.1)}$ & ${(\mathbf{93.9}, \mathbf{81.1})}$ \\
    \bottomrule
    \end{tabular}}
\end{center}
\vskip -0.2in
\caption{(Standard AUC, Robust AUC) against $l_{\infty}$-PGD attacks ($\epsilon=2/255$) on MVTec AD, obtained by AnoDDPM and ours.}
\label{table:exp_anoddpm_linf}
\vskip -0.1in
\end{table}

\subsection{Comparison with Model-agnostic Defending Strategies} \label{sec:model_agnostic}

\begin{table}[t]
\begin{center}
\resizebox{1.0\linewidth}{!}{
    \begin{tabular}{ c| c c c c c c | c }
    \toprule
       \multirow{2}{*}{Category} & \multicolumn{6}{c|}{DiffPure + CFA} & \multirow{2}{*}{AdvRAD} \\
       
       & $p=5$ & $p=25$ & $p=50$ & $p=100$ & $p=200$ & $p=300$ & \\
      \midrule
       Carpet & $9.6^{(\mathbf{96.4})}$ & $36.2^{(93.4)}$ & $51.8^{(91.9)}$ & $44.2^{(74.3)}$ & $28.6^{(45.0)}$ & $30.0^{(42.5)}$ & $\mathbf{70.5}^{(73.8)}$\\
      Grid & $24.6^{(94.0)}$ & $27.0^{(92.9)}$ & $33.3^{(85.0)}$ & $32.7^{(73.4)}$ & $21.9^{(58.2)}$ & $19.8^{(48.7)}$ & $\mathbf{99.6}^{(\mathbf{100})}$\\ 
      Leather & $47.4^{(98.3)}$ & $63.8^{(90.8)}$ & $61.6^{(86.7)}$ & $72.5^{(87.6)}$ & $57.1^{(65.9)}$ & $47.2^{(62.2)}$ & $\mathbf{97.8}^{(\mathbf{100})}$\\
    Tile & $27.4^{(\mathbf{99.4})}$ & $49.1^{(99.4)}$ & $62.5^{(97.7)}$ & $70.8^{(92.7)}$ & $67.4^{(86.7)}$ & $62.5^{(77.0)}$ & $\mathbf{93.9}^{(95.4)}$\\
      Wood & $53.7^{(\mathbf{98.9})}$ & $67.2^{(98.5)}$ & $66.8^{(92.5)}$ & $58.0^{(77.2)}$ & $48.3^{(61.1)}$ & $42.9^{(53.1)}$ & $\mathbf{95.2}^{(98.3)}$ \\
      Bottle & $93.0^{(\mathbf{99.8})}$ & $93.9^{(99.2)}$ & $95.2^{(98.9)}$ & $95.6^{(98.5)}$ & $91.0^{(96.1)}$ & $90.7^{(94.4)}$ & $\mathbf{96.9}^{(99.6)}$\\
      Cable & $3.7^{(\mathbf{89.8})}$ & $25.7^{(86.4)}$ & $38.7^{(83.3)}$ & $51.6^{(80.9)}$ & $60.3^{(79.1)}$ & $\mathbf{63.4}^{(74.6)}$ & $38.9^{(79.2)}$\\
      Capsule & $3.7^{(\mathbf{80.9})}$ & $27.0^{(70.4)}$ & $28.0^{(60.2)}$ & $31.4^{(55.1)}$ & $39.2^{(57.6)}$ & $42.0^{(54.4)}$ & $\mathbf{53.5}^{(\mathbf{90.5})}$\\
      Hazelnut & $9.8^{(\mathbf{99.8})}$ & $50.7^{(97.3)}$ & $58.4^{(94.9)}$ & $68.7^{(91.8)}$ & $77.4^{(85.4)}$ & $73.0^{(76.2)}$ & $\mathbf{91.5}^{(97.3)}$\\
      Metal Nut & $0.1^{(\mathbf{98.3})}$ & $22.0^{(90.3)}$ & $34.4^{(82.1)}$ & $37.2^{(69.7)}$ & $40.2^{(61.0)}$ & $46.6^{(59.1)}$ & $\mathbf{85.9}^{(91.0)}$\\
      Pill & $0.1^{(\mathbf{94.7})}$ & $15.2^{(89.4)}$ & $27.2^{(83.3)}$ & $38.2^{(73.4)}$ & $44.1^{(63.8)}$ & $\mathbf{44.4}^{(60.5)}$ & $39.0^{(94.4)}$\\
      Screw & $0^{(64.2)}$ & $2.8^{(55.5)}$ & $8.1^{(51.9)}$ & $15.7^{(53.7)}$ & $24.3^{(49.0)}$ & $24.4^{(43.3)}$ & $\mathbf{87.6}^{(99.3)}$\\
      Toothbrush & $32.2^{(93.6)}$ & $47.8^{(95.0)}$ & $50.6^{(87.5)}$ & $34.2^{(74.7)}$ & $9.4^{(47.5)}$ & $19.7^{(36.4)}$ & $ \mathbf{95.8}^{(\mathbf{100})} $ \\
      Transistor & $12.2^{(\mathbf{94.6})}$ & $48.6^{(92.8)}$ & $67.8^{(92.8)}$ & $75.6^{(90.4)}$ & $\mathbf{77.4}^{(87.1)}$ & $75.8^{(85.8)}$ & $74.5^{(90.6)}$\\
      Zipper & $22.7^{(95.6)}$ & $58.8^{(92.0)}$ & $69.8^{(91.5)}$ & $71.2^{(87.3)}$ & $63.6^{(80.8)}$ & $61.2^{(77.5)}$ & $\mathbf{96.2}^{(\mathbf{99.7})}$\\
      \cmidrule(rl){1-8}
      \textbf{Average} & $22.7^{(93.2)}$ & $42.4^{(89.6)}$ & $50.3^{(85.3)}$ & $53.2^{(78.7)}$ & $50.0^{(68.3)}$ & $49.6^{(63.0)}$ & $\mathbf{81.1}^{(\mathbf{93.9})}$ \\ \midrule
      $01$ & $83.1^{(100)}$ & $89.8^{(99.9)}$ & $89.7^{(99.2)}$ & $89.5^{(95.5)}$ & $89.2^{(94.6)}$ & $83.5^{(95.4)}$ & $\mathbf{100}^{(\mathbf{100})}$\\
      $02$ & $26.0^{(\mathbf{80.1})}$ & $40.6^{(77.1)}$ & $41.7^{(75.5)}$ & $46.3^{(73.6)}$ & $54.0^{(76.3)}$ & $53.5^{(71.1)}$ & $\mathbf{63.9}^{(79.9)}$\\ 
      $03$ & $16.2^{(99.5)}$ & $75.2^{(99.3)}$ & $87.0^{(98.8)}$ & $88.1^{(\mathbf{97.6})}$ & $75.5^{(88.2)}$ & $47.7^{(66.6)}$ & $\mathbf{99.5}^{(\mathbf{100})}$ \\ 
      \cmidrule(rl){1-8}
      \textbf{Average} & $41.8^{(\mathbf{93.2})}$ & $68.5^{(92.1)}$ & $72.8^{(91.2)}$ & $74.6^{(89.8)}$ & $72.9^{(86.4)}$ & $61.6^{(77.7)}$ & $\mathbf{87.8}^{(\mathbf{93.2})}$ \\
    \bottomrule
    \end{tabular}}
\end{center}
\vskip -0.2in
\caption{Robust AUC$^{\text{(Standard AUC)}}$ against $l_{\infty}$-PGD attacks on MVTec AD (first sixteen rows) and BTAD (last four rows) , obtained by  DiffPure + CFA and ours.}
\label{table:exp_diffpure}
\vskip -0.1in
\end{table}

\begin{table}[t]
\begin{center}
\resizebox{0.9\linewidth}{!}{
    \begin{tabular}{c |  c c c c  }
    \toprule
        \multirow{2}{*}{Method} & \multicolumn{4}{c}{ BTAD} \\
        & $01$  & $02$ & $03$ & \textbf{Average}\\
        \midrule
        AT + FastFlow  & $0^{(99.4)}$ & $0^{(\mathbf{85.9})}$ & $14.7^{(90.5)}$ & $4.9^{(91.9)}$ \\
        AT + CFA  & $63.4^{(98.9)}$ & $10.2^{(66.8)}$ & $5.7^{(99.6)}$ & $26.4^{(88.4)}$\\
        AdvRAD & $\mathbf{100}^{(\mathbf{100})}$ & $\mathbf{63.9}^{(79.9)}$ & $\mathbf{99.5}^{(\mathbf{100})}$ & $\mathbf{87.8}^{(\mathbf{93.2})}$\\
    \bottomrule
    \end{tabular}}
\end{center}
\vskip -0.2in
\caption{Robust AUC$^{\text{(Standard AUC)}}$ against $l_{\infty}$-PGD attacks on BTAD, obtained by  AT + FastFlow, AT + CFA and ours.}
\label{table:exp_at}
\vskip -0.2in
\end{table}

In this section, we apply two model-agnostic adversarial defenses on SOTA anomaly detectors to build defending baselines: DiffPure \citep{nie2022DiffPure} and Adversarial Training \citep{madry2018towards}, which are widely used in supervised classification. 

\noindent\textbf{Comparison with applying DiffPure on SOTA anomaly detector } We compare our method with DiffPure \citep{nie2022DiffPure}  + CFA \citep{lee2022cfa} which is the best-performing anomaly detector on clean data as presented in Table \ref{table:exp_sota_linf} and \ref{table:exp_sota_linf_btad}. We test varying purification levels $p$ (diffusion steps in DiffPure \citep{nie2022DiffPure} ) to fully verify the effectiveness of the purification strategy. Table \ref{table:exp_diffpure} summarizes the standard AUC and robust AUC against $l_{\infty}$-PGD attacks of this baseline and our method on MVTec AD and BTAD, which shows that our method enjoys a significant advantage in terms of average standard AUC and robust AUC. The results suggest that it is infeasible to tune a purification level that makes DiffPure only remove adversarial noise while preserving the anomaly signal for robust and accurate detection. See more results on ViSA in Appendix \ref{sec:more_diffpure}.

\noindent\textbf{Comparison with applying Adversarial Training on SOTA anomaly detector } We additionally compare our method with using Adversarial Training (AT) \citep{madry2018towards} on SOTA anomaly detectors. Note that AT also has flaws in protecting anomaly detectors since anomaly detection models are usually trained only on normal data \citep{bergmann2021mvtec}, which means we can only perform AT on the normal class of data without protection on the robustness of any anomalies data, thus still posing significant threats to the model. The results in Table \ref{table:exp_at} show that our method outperforms the SOTA anomaly detectors with Adversarial Training on both clean data and adversarial data. 

\subsection{Comparison with defense-enabled Anomaly Detectors} \label{sec:comp_defend}
Except for model-agnostic defending strategies, there are also domain-specific defenses explicitly designed for anomaly detection. In this section, we compare our method with APAE \citep{goodge2020robustness} and PLS \citep{lo2022adversarially}, two defense-enabled anomaly detection methods. We also compare our method with Robust Autoencoder (RAE) ~\citep{zhou2017anomaly}, which is proposed to handle noise and outlier data points. Since APAE has an optimization loop in their defense process which is hard to backpropagate, we further adopt the BPDA attack~\citep{athalye2018obfuscated} designed specifically for obfuscated gradient defenses to evaluate both our method and APAE for a fair comparison.  From Table \ref{table:exp_bpda} we can observe that \textit{AdvRAD} largely outperforms them under all attacks.

\begin{table}[htb]
\vskip -0.1in
\begin{center}
\resizebox{0.9\linewidth}{!}{
    \begin{tabular}{c c c c c c}
    \toprule
        \multirow{2}{*}{Method} & \multirow{2}{*}{Standard AUC (Avg)} & \multicolumn{4}{c}{Robust AUC (Avg)} \\
        \cmidrule(rl){3-6}
        &  & $l_{\infty}$-PGD & $l_{2}$-PGD & $l_{\infty}$-BPDA & $l_{2}$-BPDA \\
        \midrule
        RAE  & $57.1$ & $16.8$ & $49.8$ & - & - \\
        PLS  & $46.4$ & $16.0$ & $40.8$ & - & - \\
        APAE  & $64.7$ & $29.9$ & $61.2$ & $30$ & $61.2$ \\
        AdvRAD & $\mathbf{93.9}$ & $\mathbf{81.1}$ & $\mathbf{88.8}$ & $\mathbf{88.3}$ & $\mathbf{89.6}$ \\
    \bottomrule
    \end{tabular}}
\end{center}
\vskip -0.2in
\caption{Avg Robust AUC$^{\text{(Avg Standard AUC)}}$ against PGD/BPDA attacks on MVTec AD, obtained by RAE, PLS , APAE and ours.}
\label{table:exp_bpda}
\vskip -0.25in
\end{table}

\subsection{Defending against Stronger Adaptive Attacks} \label{sec:adaptive}
So far we have shown that \textit{AdvRAD} is indeed robust to PGD and BPDA attacks. To verify its robustness in more challenging settings, we test \textit{AdvRAD} against adaptive attacks where the attacker is assumed to already know about our diffusion-based anomaly detection and design attacks against our defense adaptively. Since the diffusion process in our method introduces extra stochasticity, which plays an important role in defending against adversarial perturbations, we consider applying EOT to PGD, which is designed for circumventing randomized defenses. In particular, EOT calculates the expected gradients over the randomization as a proxy for the true gradients of the inference model using Monte Carlo estimation~\citep{athalye2018synthesizing,athalye2018obfuscated,lee2022graddiv}. Table \ref{table:exp_eot} shows the robust AUC against EOT-PGD attacks on ViSA dataset. We observe that the adversarial robustness is not affected too much by EOT. Specifically, the average robust AUC slightly drops $7.4\%$ and $0.8\%$ compared against standard $l_{\infty}$-PGD and $l_{2}$-PGD attacks, respectively. These results suggest that our method has empirically strong robustness against adaptive attacks with EOT. Since other baselines use deterministic inference models, it is unnecessary to apply EOT to evaluate their adversarial robustness. Additionally, We incorporate another strong adaptive attack, AutoAttack ~\citep{croce2020reliable} which ensembles multiple white-box and black-box attacks such as APGD attacks~\citep{croce2020reliable} and Square attacks~\citep{andriushchenko2020square}. We summarize the robustness performance of method against AutoAttack in Table \ref{table:defend_AA} which we defer to Appendix \ref{sec:AA}. The robust AUC scores of \textit{AdvRAD} against AutoAttack are still largely higher than other SOTAs against relatively weaker PGD attacks.
\begin{table}[htb]
\begin{center}
\resizebox{0.9 \linewidth}{!}{
    \begin{tabular}{c | c c | c c}
    \toprule
        Category &  $l_{\infty}$-PGD & $l_{\infty}$-EOT-PGD & $l_{2}$-PGD & $l_{2}$-EOT-PGD \\
        \midrule
         PCB1 & $87.6$ & $81.3$ & $91.0$ & $90.5$ \\
         PCB2 & $94.6$ & $91.3$ & $96.2$ & $97.2$ \\
        PCB3 & $96.9$ & $92.9$  & $98.5$ &  $98.0$ \\
        PCB4 & $97.6$ & $96.0$  & $98.2$ & $97.9$ \\
         Capsules & $73.1$ & $68.7$ & $77.9$ & $76.7$\\
         Candle & $63.3$ & $46.2$ & $74.2$ & $69.8$ \\
        Macaroni1 & $83.5$ & $77.9$ & $89.5$ & $90.5$ \\
        Macaroni2 & $70.6$ & $65.6$ & $78.1$ & $77.1$ \\
         Cashew & $72.4$ & $51.2$ & $86.1$ & $86.0$ \\
         Chewing gum & $80.1$ & $67.7$ & $92.8$ & $91.1$ \\
        Fryum & $77.2$ & $72.5$ & $85.3$ & $84.8$ \\
        Pipe fryum & $79.1$ & $75.8$ & $88.8$ & $88.5$\\
        \cmidrule(rl){1-5}
        \textbf{Average} & $81.3$ & $73.9(\downarrow $7.4$)$ & $88.1$ & $87.3(\downarrow $0.8$)$ \\
    \bottomrule
    \end{tabular}}
\end{center}
\vskip -0.2in
\caption{Robust AUC against $l_{\infty}$-PGD, $l_{\infty}$-EOT-PGD, and $l_{2}$-PGD, $l_{2}$-EOT-PGD attacks (EOT=$20$) on ViSA dataset.}
\vskip -0.2in
\label{table:exp_eot}
\end{table}

\subsection{Extension: Certified Adversarial Robustness} \label{sec:rand_smoothing}

\begin{table}[t]
    \begin{center}
\resizebox{1.0\linewidth}{!}{
        \begin{tabular}{c c}
            \begin{tabular}{l r r r r}
            \toprule
                \multirow{2}{*}{Noise} & \multicolumn{4}{c}{Certified AUC at $l_2$ radius $\epsilon$} \\
                \cmidrule(rl){2-5}
                & 0 & 0.05 & 0.1 & 0.2 \\
                \midrule
                $\sigma=0.0625$ & $\mathbf{99.9}$ & $95.7$ & $0$ & $0$ \\
                $\sigma=0.125$ & $\mathbf{99.9}$ & $\mathbf{97.8}$ & $\mathbf{92.0}$ & $0$ \\
                $\sigma=0.25$ & $66.5$ & $47.3$ & $28.8$ & $\mathbf{12.4}$ \\
            \bottomrule
            \end{tabular} &
            \begin{tabular}{l r r r r}
            \toprule
                \multirow{2}{*}{Noise} & \multicolumn{4}{c}{Certified AUC at $l_2$ radius $\epsilon$} \\
                \cmidrule(rl){2-5}
                & 0 & 0.05 & 0.1 & 0.2 \\
                \midrule
                $\sigma=0.0625$ & $\mathbf{100}$ & $99.9$ & $0$ & $0$ \\
                $\sigma=0.125$ & $\mathbf{100}$ & $\mathbf{100}$ & $\mathbf{99.9}$ & $0$ \\
                $\sigma=0.25$ & $99.6$ & $99.2$ & $98.2$ & $\mathbf{98.2}$ \\
            \bottomrule
            \end{tabular} \smallskip \\
            \smallskip 
            (a) \emph{Bottle} & (b) \emph{Grid}\\
            \begin{tabular}{l r r r r}
            \toprule
                \multirow{2}{*}{Noise} & \multicolumn{4}{c}{Certified AUC at $l_2$ radius $\epsilon$} \\
                \cmidrule(rl){2-5}
                & 0 & 0.05 & 0.1 & 0.2 \\
                \midrule
                $\sigma=0.0625$ & $\mathbf{100}$ & $98.2$ & $0$ & $0$ \\
                $\sigma=0.125$ & $\mathbf{100}$ & $\mathbf{99.4}$ & $97.2$ & $0$ \\
                $\sigma=0.25$ & $\mathbf{100}$ & $\mathbf{99.4}$ & $\mathbf{98.1}$ & $\mathbf{91.7}$ \\
            \bottomrule
            \end{tabular} &
            \begin{tabular}{l r r r r}
            \toprule
                \multirow{2}{*}{Noise} & \multicolumn{4}{c}{Certified AUC at $l_2$ radius $\epsilon$} \\
                \cmidrule(rl){2-5}
                & 0 & 0.05 & 0.1 & 0.2 \\
                \midrule
                $\sigma=0.0625$ & $\mathbf{98.5}$ & $87.9$ & $0$ & $0$ \\
                $\sigma=0.125$ & $98.3$ & $\mathbf{94}$ & $\mathbf{84.8}$ & $0$ \\
                $\sigma=0.25$ & $96$ & $88.6$ & $79.2$ & $\mathbf{66.7}$ \\
            \bottomrule
            \end{tabular} \smallskip
            \\
            (c) \emph{Toothbrush} & (d) \emph{Wood} \\
        \end{tabular}}
    \end{center}
\vskip -0.25in
\caption{Certified AUC on \emph{Bottle}, \emph{Grid}, \emph{Toothbrush}, \emph{Wood} from MVTec AD at varying levels of Gaussion noise $\sigma$.}
\label{table:exp_certify}
\vskip -0.2in
\end{table}

In this section, we apply randomized smoothing ~\citep{cohen2019certified} to our diffusion-based anomaly detector and construct a new ``smoothed" detector for certified robustness.
Given a well-trained \textit{AdvRAD} detector $A_{\vtheta}(\cdot)$ that outputs the anomaly score, we can construct a binary anomaly classifier with any defined threshold $h$:
\begin{equation}
f(x)=\left\{
\begin{aligned}
& \text{normal}, \ \text{ if} \ A_{\theta}(x) \leq h \\
& \text{anomaly}, \ \text{ 
 otherwise}
\end{aligned}
\right.
\end{equation}
Then we can make predictions by constructing a Gaussian smoothed \textit{AdvRAD} and compare with $h$. The smoothed \textit{AdvRAD} enjoys provable robustness, which is summarized in the following theorem:


\begin{thm} \label{th:certify} 
[Smoothed \textit{AdvRAD}] Given a well-trained \textit{AdvRAD} detector $A_{\vtheta}(\rvx)$, for any given threshold $h$ and $\bm{\delta} \sim \mathcal{N}(0, \sigma^2\mathbf{I})$, if it satisfies $\mathbb{P}[A_{\vtheta}(\rvx+\bm{\delta}) > h] \geq p_{\text{anomaly}}(h) > 1/2$, then $\mathbb{E}_{\bm{\delta}} [A_{\vtheta}(\rvx+\bm{\delta})] > h$ for all $||\bm{\delta}||_2 < R(h)$ where $R(h) =  \sigma\Phi^{-1}(p_{\text{anomaly}}(h))$. On the other hand, if it satisfies $\mathbb{P}[A_{\vtheta}(\rvx+\bm{\delta}) < h] \geq p_{\text{normal}}(h) > 1/2$, then $\mathbb{E}_{\bm{\delta}} [A_{\vtheta}(\rvx+\bm{\delta})] < h$ for all $||\bm{\delta}||_2 < R(h)$ where $R(h) =  \sigma\Phi^{-1}(p_{\text{normal}}(h))$.
\end{thm}

Theorem \ref{th:certify} can be used to certify the robustness of a sample $\rvx$ given any threshold $h$. The estimation of $p_{\text{normal}}(h)$ and $p_{\text{anomaly}}(h)$ can be done using Monte Carlo sampling similar to~\citet{cohen2019certified}. However, the obtained certified radius is highly related to the threshold $h$. Thus the certified accuracy metric cannot fully represent the quality of the anomaly detection if the inappropriate threshold is selected. To solve this issue, we also propose the new certified AUC metric for measuring the certified robustness performance at multiple distinct thresholds. Specifically, for each threshold candidate, we can make predictions by  $\mathbb{E}_{\bm{\delta}} [A_{\vtheta}(\rvx+\bm{\delta})]$ and compute certified TPR and FPR according to prediction results and their certified radius. After iterating all possible thresholds, we calculate final AUC scores based on the collection of certified TPRs and FPRs on various thresholds. Table \ref{table:exp_certify} shows the certified robustness achieved by \textit{AdvRAD}. For example, we achieve $98.2\%$ certified AUC at $l_2$ radius $0.2$ on \emph{gird} sub-dataset, which indicates that there does not exist any adversarial perturbations $\bm{\delta}$ ($|| \bm{\delta} || \leq 0.2$) that can make the AUC lower than $98.2\%$. One major limitation of randomized smoothing on anomaly detection tasks is that the noise level can not be much high, otherwise the anomalous features might be covered by the Gaussian noise such that the detector can not distinguish anomalous samples from normal samples.
\section{Conclusion}
Adversarial robustness is a critical factor for the practical deployment of 
industrial anomaly detection models. In this work, we first identify that naively applying the state-of-the-art empirical defense, adversarial purification, to anomaly detection suffers from a high anomaly miss rate as the purifier can also purify the anomaly signals along with the adversarial perturbations. We further propose \textit{AdvRAD} based on diffusion models to perform anomaly detection and adversarial purification simultaneously.  We leverage extensive evaluation to validate that \textit{AdvRAD} outperforms existing SOTA methods by a significant margin in adversarial robustness.

{
    \small
    \bibliographystyle{ieeenat_fullname}
    \bibliography{main}
}
\clearpage
\appendix
\section{Training Objective of the Diffusion Model} \label{sec:loss}
In this section, we introduce the hybrid training objective proposed by \citet{nichol2021improved}. Specifically, training diffusion models can be performed by optimizing the commonly used variational bound on negative log-likelihood as follows~\citep{ho2020denoising}: 
\begin{align}
L_{\text{vb}} &:= L_{0} + L_{1} + \ldots + L_{T-1} + L_{T}\\
L_{0} &:= -\log p_\vtheta(\rvx_0|\rvx_1)\\
L_{t-1} &:= D_{KL}(q(\rvx_{t-1}|\rvx_t,\rvx_0)||p_\vtheta(\rvx_{t-1}|\rvx_t)) \label{eq:posterior} \\
L_{T} &:= D_{KL}(q(\rvx_T|\rvx_0)||p(\rvx_T))
\end{align}

\citet{ho2020denoising} suggest that directly optimizing this variational bound $L_{vb}$ would produce much more gradient noise during training and propose a reweighted simplified objective $L_{simple}$:
\begin{equation}
    L_{\text{simple}} = \E_{t, \rvx_0, \bm{\epsilon}}[\| \bm{\epsilon} - \bm{\epsilon}_{\vtheta}(\rvx_{t}, t) \|].
\end{equation}
However, this $L_{simple}$ model suffers from sample quality loss when using a reduced number of denoising steps~\citep{nichol2021improved}. \citet{nichol2021improved} find that training diffusion models via a hybrid objective:
\begin{equation}
    L_{\text{hybrid}} = L_{\text{simple}} + \lambda L_{\text{vb}}
\end{equation}
greatly improves its practical applicability by generating high-quality samples with fewer denoising steps, which is helpful for using diffusion models on applications with high-efficiency requirements such as real-time anomaly detection~\citep{sun2021ctf}. In particular, we parameterize the variance term $\boldsymbol{\Sigma}_\vtheta(\rvx_t, t)$ as an interpolation between $\beta_t$ and $\tilde{\beta}_t$ in the log domain following~\citep{nichol2021improved}:
\begin{equation}
    \boldsymbol{\Sigma}_\vtheta(\rvx_t, t) = exp(\rvv\log \beta_t+(1-\rvv)\log \tilde{\beta}_t)
\end{equation}
where $\rvv$ is the model output. Following \citet{nichol2021improved}, we set $\lambda=0.001$ and apply a stop-gradient to the $\boldsymbol{\mu}_\vtheta(\rvx_t,t)$ output for $L_{\text{vb}}$ to prevent $L_{\text{vb}}$ from overwhelming $L_{\text{simple}}$. The hybrid objective can allow fewer denoising steps while maintaining high-quality generation \citep{nichol2021improved}, which gives us the opportunity to explore different denoising steps for the anomaly detection task. See the ablation study of the impact of denoising steps in Appendix \ref{sec:fewer_shot}.

\section{Additional Algorithms} \label{sec:full-shot}
\subsection{Full-shot Robust Reconstruction in \textit{AdvRAD}}
\begin{algorithm}[htb]
   \caption{Full-shot Robust Reconstruction in \textit{AdvRAD}}
   \label{alg:full-shot}
\begin{algorithmic}[1]
   \STATE {\bfseries Input:} Test images: $\rvx$, diffusion steps: $k(k\leq T)$
   \STATE {\bfseries Output:} Reconstructions of $\rvx$: $\tilde{\rvx}$
   \STATE $\rvx_0 = \rvx$
   \STATE $\bm{\epsilon} \sim \mathcal{N}(0, \mathbf{I})$
   \STATE $\rvx_k = \sqrt{\overline{\alpha}_k}\rvx_0 + \sqrt{1-\overline{\alpha}_k}\bm{\epsilon}$ \\ //  full-shot denoising process:
   \FOR{$t=k$ {\bfseries to} $1$}
   \STATE $\tilde{\rvx}_0 = \frac{1}{\sqrt{\overline{\alpha}_{t}}}({\rvx}_{t}- \sqrt{1-\overline{\alpha}_{t}}\bm{\epsilon}_{\vtheta}(\rvx_{t}, t))$
   \IF{$t>1$}
   \STATE $\rvz \sim \mathcal{N}(0, \mathbf{I})$
   \STATE $\rvx_{{t-1}} = \frac{\sqrt{\overline{\alpha}_{{t-1}}}\beta_{t}}{1-\overline{\alpha}_{t}} \tilde{\rvx}_0+\frac{\sqrt{\overline{\alpha}_{t}}(1-\overline{\alpha}_{t-1})}{1-\overline{\alpha}_{t}} \rvx_{t} + \sqrt{\bm{\Sigma}_{\vtheta}(\rvx_{t}, t))}\rvz$
   \ENDIF
   \ENDFOR
   \STATE $\tilde{\rvx} = \tilde{\rvx}_0$
\end{algorithmic}
\end{algorithm}

Algorithm \ref{alg:full-shot} summarizes the main steps for full-shot robust reconstruction. Specifically, we first choose the diffusion steps $k$ and apply Eq. \ref{eq:diffusion} on $\rvx$ to obtain diffused images $\rvx_k$.
Unlike the diffusion model training process, here we do not need to diffuse the data into complete Gaussian noise (a large $k$). Instead, we pick a moderate number of $k$ for noise injection and start denoising thereafter. The key difference between the robust reconstruction in anomaly detection and the purification process in DiffPure \citep{nie2022DiffPure} is that $k$ should be chosen such that the amount of Gaussian noise is dominating the anomaly signals and adversarial perturbations while the high-level features of the input data are still preserved for reconstruction. In terms of the denoising process, a typical full-shot setting uses the full $k$ denoising steps: in each step $t$, we iteratively predict the true input $\rvx$ given the current diffused data $\rvx_{t}$, termed $\tilde{\rvx}_0$, then sampling the new iterate $\rvx_{t-1}$ according to the current prediction $\tilde{\rvx}_0$ and the current diffused data $\rvx_{t}$. 

\subsection{Arbitrary-shot Robust Reconstruction in \textit{AdvRAD}} 
We attach the complete algorithm for arbitrary-shot robust reconstruction motivated by \citet{nichol2021improved} in \cref{alg:arbitrary-shot}. Given an arbitrary denoising steps $S=\{S_m,S_{m-1}, \dots, S_1\} (m \leq k, k=S_m>S_{m-1}>\dots >S_{1}>=1)$, in each step $t \in [1, m]$,
we iteratively predict the true point $\rvx$ given the current diffused data $\rvx_{S_t}$, 
termed $\tilde{\rvx}_0$, them sampling new iterate $\rvx_{S_{t-1}}$ according to the current prediction $\tilde{\rvx}_0$ and current diffused data $\rvx_{S_t}$.
\label{sec:arbitrary-shot}

\begin{algorithm}[htb]
   \caption{Arbitrary-shot Robust Reconstruction in \textit{AdvRAD}}
   \label{alg:arbitrary-shot}
\begin{algorithmic}[1]
   \STATE {\bfseries Input:} Test images: $\rvx$, diffusion steps: $k$, arbitrary generation steps: $S=\{S_m,S_{m-1}, \dots, S_1\} (m \leq k, k=S_m>S_{m-1}>\dots >S_{1}>=1)$
   \STATE {\bfseries Output:} Reconstructions of $\rvx$: $\tilde{\rvx}$
   \STATE $\rvx_0 = \rvx$
   \STATE $\bm{\epsilon} \sim \mathcal{N}(0, \mathbf{I})$
   \STATE $\rvx_k = \sqrt{\overline{\alpha}_k}\rvx_0 + \sqrt{1-\overline{\alpha}_k}\bm{\epsilon}$ \\ //  arbitrary-shot denoising process:
   \FOR{$t=m$ {\bfseries to} $1$}
   \STATE $\tilde{\rvx}_0 = \frac{1}{\sqrt{\overline{\alpha}_{S_t}}}({\rvx}_{S_t}- \sqrt{1-\overline{\alpha}_{S_t}}\bm{\epsilon}_{\vtheta}(\rvx_{S_t}, S_t))$
   \IF{$t>1$}
   \STATE $\rvz \sim \mathcal{N}(0, \mathbf{I})$
   \STATE $\rvx_{S_{t-1}} = \frac{\sqrt{\overline{\alpha}_{S_{t-1}}}\beta_{S_t}}{1-\overline{\alpha}_{S_t}} \tilde{\rvx}_0+\frac{\sqrt{\overline{\alpha}_{S_t}}(1-\overline{\alpha}_{S_{t-1}})}{1-\overline{\alpha}_{S_t}} \rvx_{S_t} + \sqrt{\bm{\Sigma}_{\vtheta}(\rvx_{S_t}, S_t))}\rvz$
   \ENDIF
   \ENDFOR
   \STATE $\tilde{\rvx} = \tilde{\rvx}_0$
\end{algorithmic}
\end{algorithm}

\subsection{Anomaly Score Calculation} \label{sec:score}

We attach the complete algorithm for anomaly score calculation in \cref{alg:score}. Given
test image $\rvx \in \R^{C \times H \times W}$ and its reconstruction $\tilde{\rvx} \in \R^{C \times H \times W}$ obtained by \textit{AdvRAD}, we first calculate the Multiscale Reconstruction Error Map. In particular, we choose a scale schedule $L=\{1, \frac{1}{2}, \frac{1}{4}, \frac{1}{8}\}$. For each scale $l$, we compute the error map ${\text{Err}(\rvx, \tilde{\rvx})}_l$ between the downsampled input ${\rvx}^l$ and the downsampled reconstruction ${\tilde{\rvx}}^l$ with  $\frac{1}{C}\sum_{c=1}^{C}{({\rvx}^l-{\tilde{\rvx}}^l)}_{[c,:,:]}^2$ where the square operator here refers to element-wise square operation, then unsampled to the original resolution. The final $\text{Err}_{\text{ms}}$ is obtained by averaging each scale's error map and applying a mean filter for better stability similar to \citet{zavrtanik2021reconstruction}: ${\text{Err}_{\text{ms}}(\rvx, \tilde{\rvx})} = (\frac{1}{N_L}\sum_{l \in L}{\text{Err}(\rvx, \tilde{\rvx})}_l) \ast f_{s \times s}$
where $f_{s \times s}$ is the mean filter of size ${s \times s}$, $\ast$ is the convolution operation. Similar to \citet{pirnay2022inpainting}, we take the pixel-wise maximum of the absolute deviation of the $\text{Err}_{\text{ms}}(\rvx, \tilde{\rvx})$ to the normal training data as the scalar anomaly score. 
\begin{algorithm}[htb]
   \caption{Anomaly Score Calculation in \textit{AdvRAD}}
   \label{alg:score}
\begin{algorithmic}[1]
   \STATE {\bfseries Input:} Test image: $\rvx \in \R^{C \times H \times W}$, Reconstructed image: $\tilde{\rvx} \in \R^{C \times H \times W}$
   \STATE {\bfseries Output:} Anomaly score: $A(\rvx)$ \\ //$L$ is a downsampling scale schedule:
   \FOR{$l$ in $L=\{1, \frac{1}{2}, \frac{1}{4}, \frac{1}{8}\}$}
   \STATE ${\rvx}^l = \text{downsample}(l, \rvx) \in \R^{C \times (l \times H) \times (l \times W)}$
   \STATE ${\tilde{\rvx}}^l = \text{downsample}(l, \tilde{\rvx}) \in \R^{C \times (l \times H) \times (l \times W)}$ \\ //element-wise sqaure:
   \STATE ${\text{Err}(\rvx, \tilde{\rvx})}_l=\text{upsample}(\frac{1}{l}, \frac{1}{C}\sum_{c=1}^{C}{({\rvx}^l-{\tilde{\rvx}}^l)}_{[c,:,:]}^2) \in \R^{H \times W}$
   \ENDFOR \\ //$f_{s \times s}$ is a mean filter of size $(s \times s)$:
   \STATE ${\text{Err}_{\text{ms}}(\rvx, \tilde{\rvx})} = (\frac{1}{N_L}\sum_{l \in L}{\text{Err}(\rvx, \tilde{\rvx})}_l) \ast f_{s \times s} \in \R^{H \times W}$ \\ //$Z$ is the set of normal training images:
   \STATE $A(\rvx) = \text{max}({|\text{Err}_{\text{ms}}(\rvx, \tilde{\rvx})- \frac{1}{N_Z}\sum_{z \in Z}\text{Err}_{\text{ms}}(\rvz, \tilde{\rvz})|})$ 
\end{algorithmic}
\end{algorithm}

\section{More Details of the Experimental Settings}
\subsection{Hyperparameters of the Diffusion Model} \label{sec:hyper}
The diffusion model in our experiments uses the linear noise schedule~\citep{ho2020denoising} by default. The number of channels in the first layer is 128, and the number of heads is 1. The attention resolution is $16 \times 16$. We adopt PyTorch as the deep learning framework for implementations. We train the model using Adam optimizer with the learning rate of $10^{-4}$ and the batch size 2. The model is trained for 30000 iterations for all categories of MVTec AD and 120000 iterations for all categories of ViSA. We set diffusion steps $T=1000$ for training. We set diffusion step $k=100$ at inference time for all categories of data. See more ablation studies on hyperparameters in Appendix \ref{sec:k} and \ref{sec:training_hyper}.

\section{More Experimental Results}

\subsection{Comparison with the SOTA Anomaly Detectors against $l_2$ Bounded Attacks} \label{sec:l2}
As shown in Table \ref{table:exp_sota_l2}, we summarize the robustness performance of different SOTA anomaly detectors and ours against $l_2$-PGD attacks ($\epsilon=0.2$) on the MVTec AD dataset. We can observe that our method improves average robust AUC against $l_{2}$-PGD attacks ($\epsilon=0.2$) by $44.9\%$ and achieves $88.8\%$ robust AUC, which indicates that our method is still robust to $l_2$ bounded PGD attacks.

\begin{table}[htb]
\begin{center}
\resizebox{1.0\linewidth}{!}{
    \begin{tabular}{ c| c c c c c | c c }
    \toprule
       Category & OCR-GAN  & SPADE  & CFlow  & FastFlow  & CFA & AdvRAD \\
      \midrule
       Carpet & $18.5^{(76.6)}$ & $27.1^{(92.8)}$ & $13.5^{(98.6)}$ & $18^{(\mathbf{99.7})}$ & $65.1^{(99.4)}$ & $\mathbf{72.9}^{(73.8)}$ \\
      Grid & $0^{(97)}$ & $4.1^{(47.3)}$ & $0^{(96.6)}$ & $0^{(\mathbf{100})}$ & $50^{(99.6)}$ & $\mathbf{99.9}^{(\mathbf{100})}$ \\ 
      Leather & $0^{(90.7)}$ & $16.5^{(95.4)}$ & $9.4^{(\mathbf{100})}$ & $35.4^{(\mathbf{100})}$ & $77.6^{(\mathbf{100})}$ & $\mathbf{99.9}^{(\mathbf{100})}$ \\
      Tile & $7.4^{(95.6)}$ & $45.9^{(96.5)}$ & $7.8^{(99.6)}$ & $30.5^{(\mathbf{100})}$ & $72.4^{(99.3)}$ & $\mathbf{94.8}^{(95.4)}$ \\
      Wood & $0^{(95.4)}$ & $11^{(95.8)}$ & $18.1^{(99.7)}$ & $22^{(\mathbf{99.9})}$ & $61.8^{(99.7)}$ & $\mathbf{95.5}^{(98.3)}$ \\
      \midrule
       Bottle & $0.1^{(97.7)}$ & $0^{(97.2)}$ & $48.5^{(\mathbf{100})}$ & $2.2^{(\mathbf{100})}$ & $74.6^{(\mathbf{100})}$ & $\mathbf{97.5}^{(99.6)}$ \\
      Cable & $3.2^{(71.5)}$ & $0.9^{84.8)}$ & $19.2^{(98.7)}$ & $0.3^{(67.4)}$ & $\mathbf{69.5}^{(\mathbf{99.8})}$ & $65.7^{(79.2)}$ \\ 
      Capsule & $0^{(80.4)}$ & $0^{(89.7)}$ & $1.6^{(93.7)}$ & $13.8^{(\mathbf{99.2})}$ & $1.7^{(97)}$ & $\mathbf{68.1}^{(90.5)}$ \\ 
      Hazelnut & $18.5^{(97.7)}$ & $0^{(88.1)}$ & $4.9^{(99.9)}$ & $0.8^{(99.5)}$ & $47.2^{(\mathbf{100})}$ & $\mathbf{94.3}^{(97.3)}$ \\
      Metal Nut & $2.8^{(82.6)}$ & $0^{(71)}$ & $4.4^{(\mathbf{100})}$ & $1.7^{(98.2)}$ & $14.3^{(\mathbf{100})}$ & $\mathbf{87.9}^{(91.0)}$ \\
      Pill & $2.7^{(80.8)}$ & $0.4^{(80.1)}$ & $0^{(93.2)}$ & $0^{(97.8)}$ & $3.3^{(\mathbf{98})}$ & $\mathbf{80.3}^{(94.4)}$ \\ 
      Screw & $0^{(\mathbf{99.4)}}$ & $0^{(66.7)}$ & $0^{(79)}$ & $6.6^{(91.1)}$ & $0^{(95.5)}$ & $\mathbf{91.8}^{(99.3)}$ \\ 
      Toothbrush & $0^{(96.7)}$ & $0^{(88.9)}$ & $18.3^{(85.3)}$ & $3.6^{(94.7)}$ & $38.3^{(\mathbf{100})}$ & $\mathbf{99.4}^{(\mathbf{100})}$ \\
      Transistor & $1.7^{(75)}$ & $4.8^{(90.3)}$ & $8.8^{(98.3)}$ & $0.4^{(99.4})$ & $53.7^{(\mathbf{100)}}$ & $\mathbf{84.3}^{(90.6)}$ \\
      Zipper & $0^{(80.4)}$ & $3.2^{(96.6)}$ & $0^{(97.5)}$ & $19.3^{(99.6)}$ & $29.2^{(99.7)}$ & $\mathbf{99.2}^{(\mathbf{99.7})}$ \\
      \midrule
      \textbf{Average} & $3.7^{(87.8)}$ & $7.59^{(85.4)}$ & $10.3^{(96.0)}$ & $9.9^{(98.5)}$ & $43.9^{(\mathbf{99.2})}$ & $\mathbf{88.8}^{(93.9)}$ \\
    \bottomrule
    \end{tabular}}
\end{center}
\vskip -0.1in
\caption{Standard AUC (in parenthesis) and robust AUC against $l_{2}$-PGD attacks ($\epsilon=0.2$) on MVTec AD dataset, obtained by different state-of-the-art anomaly detectors and ours.}
\vskip -0.1in
\label{table:exp_sota_l2}
\end{table}

\subsection{More Results of the Comparison with applying DiffPure on SOTA anomaly detector} 
\label{sec:more_diffpure}

Here we additionally compare with DiffPure \citep{nie2022DiffPure}  + CFA \citep{lee2022cfa} by using the PGD attack with $l_\infty$ perturbations on ViSA dataset. The results are shown in Table \ref{table:exp_diffpure_visa}. We can see that our method still largely outperforms this baseline, with an absolute improvement of $+41.3\%$ in robust AUC and an absolute improvement of $+7.3\%$ in standard AUC.

\begin{table}[t]
\begin{center}
\resizebox{1.0\linewidth}{!}{
    \begin{tabular}{ c| c c c c c c | c }
    \toprule
       \multirow{2}{*}{Category} & \multicolumn{6}{c|}{DiffPure + CFA} & \multirow{2}{*}{AdvRAD} \\
       & $p=5$ & $p=25$ & $p=50$ & $p=100$ & $p=200$ & $p=300$ & \\
      \midrule
       PCB1 & $5.9^{(93.0)}$ & $54.3^{(91.9)}$ & $70.0^{(91.1)}$ & $51.8^{(78.0)}$ & $36.2^{(60.5)}$ & $34.4^{(54.2)}$ & $\mathbf{87.6}^{(\mathbf{95.9})}$\\
      PCB2 & $6.2^{(93.2)}$ & $37.3^{(90.9)}$ & $52.2^{(86.8)}$ & $44.1^{(71.9)}$ & $24.5^{(34.7)}$ & $21.5^{(33.0)}$ & $\mathbf{94.6}^{(\mathbf{99.4})}$\\ 
      PCB3 & $15.5^{(93.6)}$ & $36.4^{(89.3)}$ & $47.2^{(85.0)}$ & $49.8^{(77.0)}$ & $43.5^{(63.9)}$ & $46.8^{(64.1)}$ & $\mathbf{96.9}^{(\mathbf{99.4})}$\\
    PCB4 & $2.2^{(\mathbf{99.6})}$ & $45.7^{(99.3)}$ & $45.1^{(81.7)}$ & $23.0^{(57.3)}$ & $17.2^{(40.8)}$ & $21.9^{(35.0)}$ & $\mathbf{97.6}^{(99.3)}$\\
      Capsules & $0.5^{(64.3)}$ & $17.3^{(63.9)}$ & $28.4^{(63.0)}$ & $26.0^{(58.1)}$ & $27.6^{(52.3)}$ & $35.3^{(47.9)}$ & $\mathbf{73.1}^{(\mathbf{88.4})}$ \\
      Candle & $2.4^{(90.9)}$ & $8.6^{(87.3)}$ & $15.3^{(85.1)}$ & $35.5^{(87.3)}$ & $53.5^{(80.5)}$ & $46.4^{(68.2)}$ & $\mathbf{63.3}^{(\mathbf{92.2})}$\\
      Macaroni1 & $1.8^{(83.8)}$ & $19.3^{(73.8)}$ & $21.3^{(67.4)}$ & $12.1^{(51.9)}$ & $21.3^{(45.5)}$ & $48.6^{(54.2)}$ & $\mathbf{83.5}^{(\mathbf{99.0})}$\\
      Macaroni2 & $0^{(69.4)}$ & $3.2^{(65.8)}$ & $5.0^{(62.9)}$ & $2.8^{(58.1)}$ & $13.3^{(60.3)}$ & $35.2^{(56.1)}$ & $\mathbf{70.6}^{(\mathbf{93.0})}$\\
      Cashew & $3.0^{(93.8)}$ & $29.9^{(91.1)}$ & $39.5^{(87.1)}$ & $37.9^{(75.8)}$ & $35.6^{(64.5)}$ & $30.4^{(41.1)}$ & $\mathbf{72.4}^{(\mathbf{97.3})}$\\
      Chewing gum & $5.0^{(\mathbf{99.4})}$ & $47.9^{(95.9)}$ & $56.0^{(90.9)}$ & $50.1^{(75.9)}$ & $40.7^{(66.1)}$ & $47.9^{(64.0)}$ & $\mathbf{80.1}^{(\mathbf{99.4})}$\\
      Fryum & $5.7^{(91.3)}$ & $47.1^{(88.6)}$ & $54.0^{(84.5)}$ & $58.4^{(80.0)}$ & $61.1^{(76.9)}$ & $63.4^{(75.2)}$ & $\mathbf{77.2}^{(\mathbf{93.7})}$\\
      Pipe fryum & $2.1^{(95.3)}$ & $23.1^{(72.3)}$ & $21.8^{(67.0)}$ & $19.9^{(76.3)}$ & $28.8^{(78.5)}$ & $47.9^{(79.3)}$ & $\mathbf{79.1}^{(\mathbf{98.3})}$\\
      \cmidrule(rl){1-8}
      \textbf{Average} & $4.2^{(89.0)}$ & $30.8^{(84.2)}$ & $38.0^{(79.4)}$ & $34.3^{(70.6)}$ & $33.6^{(60.4)}$ & $40.0^{(56.0)}$ & $\mathbf{81.3}^{(\mathbf{96.3})}$ \\
    \bottomrule
    \end{tabular}}
\end{center}
\vskip -0.1in
\caption{Robust AUC$^{\text{(Standard AUC)}}$ against $l_{\infty}$-PGD attacks on ViSA, obtained by  DiffPure + CFA and ours.}
\label{table:exp_diffpure_visa}
\vskip -0.15in
\end{table}

\subsection{Defending against AutoAttack} \label{sec:AA}
In this section, we incorporate additional strong attack baselines, AutoAttack ~\citep{croce2020reliable} which ensemble multiple white-box and black-box attacks such as APGD attacks and Square attacks. Specifically, we used two versions of AutoAttack: (i) standard AutoAttack and (ii) random AutoAttack (EOT+AutoAttack), which is used for evaluating stochastic defense methods. We summarize the standard AUC and robust AUC of our proposed \textit{AdvRAD} in the following Table \ref{table:defend_AA}. The robust AUC scores of \textit{AdvRAD} against AutoAttack are still largely higher than other SOTAs against relatively weaker PGD attacks as shown in Table \ref{table:exp_sota_linf}, \ref{table:exp_sota_l2}, and \ref{table:details_comp_defense_enabled}. Thus, there is no need to evaluate other methods' robustness against stronger AutoAttack.
\begin{table}[htb]
\begin{center}
\resizebox{1.0\linewidth}{!}{
    \begin{tabular}{c | c c c c c}
    \toprule
        \multirow{2}{*}{Category} & \multirow{2}{*}{\makecell[c]{Standard \\ AUC}} & \multicolumn{4}{c}{Robust AUC} \\
        \cmidrule(rl){3-6}
        &  & $l_{\infty}$-standard AA & $l_{2}$-standard AA & $l_{\infty}$-random AA & $l_{2}$-random AA \\
        \midrule
         Bottle & $99.6$ & $76.5$ & $87.8$ & $73.4$ & $87.6$\\
        Grid & $100$ & $98.2$ & $99.2$ & $98.2$ & $98.8$\\
        Toothbrush & $100$ & $73.6$ & $84.2$ & $65.8$ & $86.1$\\
        wood & $98.3$ & $72.3$ & $75.2$ & $64.8$ & $75.3$\\
        \cmidrule(rl){1-6}
        \textbf{Average} & $99.6$ & $80.1$ & $86.6$ & $75.6$ & $87.0$\\
    \bottomrule
    \end{tabular}}
\end{center}
\vskip -0.1in
\caption{Standard AUC and robust AUC against $l_{\infty}$-AutoAttack($\epsilon=2/255$), $l_{2}$-AutoAttack($\epsilon=0.2$) on \emph{Bottle}, \emph{Grid}, \emph{Toothbrush}, \emph{Wood} from MVTec AD}
\vskip -0.1in
\label{table:defend_AA}
\end{table}

\subsection{Impact of the Diffusion Step} \label{sec:k}
Here we first provide the anomaly detection performance of our proposed \textit{AdvRAD} on clean data with varying diffusion steps $k$ at inference time. We test with $t \in \{50, 100, 200, 300 \}$ on MVTec AD dataset. As shown in Table \ref{table:k_clean_ablation}, different categories may have different optimal $k$. While the diffusion step $k$ does impact the performance on individual categories, we observe a stable performance over a range of k, dropping only at $k=300$. In terms of the adversarial data, the robust AUC against $l_{\infty}$-PGD attacks ($\epsilon=2/255$) for varying $k$ are shown in Table \ref{table:k_rob_ablation}. We observe that when $k \leq 200$, \textit{AdvRAD} obtain better robustness with higher $k$. The robust AUC at $k = 300$ slightly
decreases compared with $k = 200$, which is due to the impact of the performance decrease on clean data.

\begin{table}[tb]
\begin{center}
\resizebox{0.85\linewidth}{!}{
    \begin{tabular}{ c | c c c c }
    \toprule
      Category  & $k=50$ & $k=100$ & $k=200$ & $k=300$ \\
      \midrule
       Carpet  & $64.9$ & $73.8$ & $\mathbf{82.7}$ & $80.9$ \\
      Grid  & $\mathbf{100}$ & $\mathbf{100}$ & $\mathbf{100}$ & $\mathbf{100}$\\ 
      Leather  & $\mathbf{100}$ & $\mathbf{100}$ & $99.3$ & $98.4$ \\
      Tile  & $\mathbf{99.2}$ & $95.4$ & $81.4$ & $74.0$ \\
      Wood & $98.2$ & $\mathbf{98.3}$ & $97.9$ & $97.1$\\\midrule
      Bottle  & $\mathbf{100}$ & $99.6$ & $99.1$ & $97.9$\\
      Cable  & $78.8$ & $79.2$ & $\mathbf{79.5}$ & $77.7$\\ 
      Capsule  & $\mathbf{93.9}$ & $90.5$ & $84.6$ & $80.7$\\ 
      Hazelnut  & $96.2$ & $97.3$ & $\mathbf{97.5}$ & $96.2$\\
      Metal Nut & $83.8$ & $91.0$ & $91.3$ & $\mathbf{93.5}$\\
      Pill  & $\mathbf{97.2}$ & $94.4$ & $86.6$ & $68.6$ \\
      Screw  & $95.0$ & $\mathbf{99.3}$ & $80.8$ & $66$\\ 
      Toothbrush  & $\mathbf{100}$ & $\mathbf{100}$ & $99.7$ & $99.7$\\
      Transistor  & $87.8$ & $90.6$ & $\mathbf{93.7}$ & $93.2$ \\
      Zipper & $\mathbf{100}$ & $99.7$ & $96.4$ & $95.0$\\
      \midrule
      Average & $93.0$ & $\mathbf{93.9}$ & $91.4$ & $87.9$\\
    \bottomrule
    \end{tabular}}
\end{center}
\vskip -0.1in
\caption{Standars AUC results on 15 categories from MVTec AD with varying diffusion steps $k$ at inference time}
\label{table:k_clean_ablation}
\vskip -0.1in
\end{table}

\begin{table}[tb]
\begin{center}
\resizebox{0.8\linewidth}{!}{
    \begin{tabular}{ c | c c c c }
    \toprule
      Category  & $k=50$ & $k=100$ & $k=200$ & $k=300$ \\
      \midrule
      Bottle  & $88.6$ & $96.9$ & $97.1$ & $\mathbf{97.5}$ \\
      Grid  & $99.2$ & $99.6$ & $\mathbf{99.7}$ & $99.4$\\ 
      Toothbrush  & $93.6$ & $95.8$ & $\mathbf{97.5}$ & $\mathbf{97.5}$ \\
      Wood & $85.4$ & $\mathbf{95.2}$ & $95.1$ & $92.9$\\\midrule
      Average & $91.7$ & $94.7$ & $\mathbf{97.3}$ & $96.8$\\
    \bottomrule
    \end{tabular}}
\end{center}
\vskip -0.15in
\caption{Robust AUC results against $l_{\infty}$-PGD attacks ($\epsilon=2/255$) on \emph{Bottle}, \emph{Grid}, \emph{Toothbrush}, \emph{Wood}  from MVTec AD with varying diffusion steps $k$ at inference time}
\label{table:k_rob_ablation}
\vskip -0.2in
\end{table}

\subsection{Reducing the Denoising Steps} \label{sec:fewer_shot}
In this section, we provide the anomaly detection performance of \textit{AdvRAD} on clean data at varying denoising steps as shown in Table \ref{table:fewer-shot} by running Algorithm \ref{alg:arbitrary-shot} for reconstruction and using Algorithm \ref{alg:score} to compute anomaly score. Specifically, we test with several denoising steps schedules from one-shot denoising ($1$-step) to full-shot denoising ($k$-step) and intermediate settings such as $0.05k$, $0.1k$, $0.25k$, and $0.5k$. We can see that one-shot denoising obtains the highest AUC scores on all three categories. Moreover, we report the inference time (in seconds) at varying denoising steps in Table \ref{table:inferece_time} on an NVIDIA TESLA K80 GPU, where the inference time increases linearly with denoising steps. We show that the inference with one-shot denoising could process a single image in 0.5 seconds, which demonstrates the applicability of our method \textit{AdvRAD} on real-time tasks. These experimental results clearly indicate that \textit{AdvRAD} with reconstruction by one-shot denoising achieves both the best detection effectiveness and time efficiency.
\begin{table}[htb]
\begin{center}
\resizebox{1.0\linewidth}{!}{
    \begin{tabular}{ c | c c c c c c }
    \toprule
      Category & $1-$step & $0.05k$-step & $0.1k$-step & $0.25k$-step & $0.5k$-step & $k$-step \\
      \midrule
      Screw & $\mathbf{99.3}$ & $97.3$ & $96.8$ & $97.2$ & $95.1$ & $96.4$ \\ 
      Toothbrush & $\mathbf{100}$ & $99.7$ & $\mathbf{100}$ & $99.4$ & $\mathbf{100}$ & $\mathbf{100}$ \\
      Wood & $\mathbf{98.3}$ & $97.1$ & $98.2$ & $95.9$ & $98.2$ & $96$ \\ 
    \bottomrule
    \end{tabular}}
\end{center}
\vskip -0.1in 
\caption{AUC results on \emph{Screw}, \emph{Toothbrush}, \emph{Wood} at varying denoising steps.}
\vskip -0.1in 
\label{table:fewer-shot}
\end{table}

\begin{table}[htb]
\begin{center}
\resizebox{1.0\linewidth}{!}{
    \begin{tabular}{ c | c c c c c c }
    \toprule
      Category & $1$-step & $0.05k$-step & $0.1k$-step & $0.25k$-step & $0.5k$-step & $k$-step \\
      \midrule
      Toothbrush & $0.5$ & $2.28_{(\times 4.6)}$ & $4.63_{(\times 9.3)}$ & $11.53_{(\times 23.6)}$ & $23.03_{(\times 46.1)}$ & $46.06_{(\times 92.1)}$ \\
    \bottomrule
    \end{tabular}}
\end{center}
\vskip -0.1in 
\caption{Inference time (in seconds) for a single image on \emph{Toothbrush} by varying denoising steps, where the inference time increases over one-shot denoising is given in parenthesis.}
\label{table:inferece_time}
\vskip -0.2in 
\end{table}

\subsection{Comparison with Gaussian-noise Injection Defense}
In this section, we test the defense strategy of applying Gaussian-noise injection as the data augmentation. To verify it, we train two SOTA anomaly detectors FastFlow \citep{yu2021fastflow} and CFA \citep{lee2022cfa} with Gaussian-noise Injection (GN) to evaluate their performance. Specifically, in the training process, we randomly inject Gaussian noise with the varying standard deviation from $0, 0.005, 0.01, 0.05,0.1$. Note that the $l_\infty$ attacks in our work are bounded in $2/255$ thus the injected Gaussian noise can ``dominate'' the adversarial perturbation. We present the standard AUC (in parenthesis) and robust AUC against $l_\infty$-PGD attacks in Table \ref{table:gn}. We can clearly observe that adding Gaussian-noised samples in other baselines can not improve their adversarial robustness.

\begin{table}[htb]
\begin{center}
\resizebox{0.8\linewidth}{!}{
    \begin{tabular}{ c | c c c }
    \toprule
      Category  & FastFlow +GN & CFA + GN & AdvRAD \\
      \midrule
      Bottle  & $1.4^{(\mathbf{100})}$ & $10.2^{(\mathbf{100})}$ & $\mathbf{96.9}^{(99.6)}$ \\
      Grid  & $0^{(97.4)}$ & $0^{(\mathbf{100})}$ & $\mathbf{99.6}^{(\mathbf{100})}$ \\
      Toothbrush  & $0^{(99.7)}$ & $0^{(91.9)}$ & $\mathbf{95.8}^{(\mathbf{100})}$ \\
      Wood & $0.4^{(\mathbf{99.6})}$ & $0.6^{(99.2)}$ & $\mathbf{95.2}^{(98.3)}$ \\
      \midrule
      Average & $0.45^{99.2}$ & $2.7^{97.8}$ & $\mathbf{96.9}^{(\mathbf{99.6})}$ \\
    \bottomrule
    \end{tabular}}
\end{center}
\vskip -0.1in
\caption{Robust AUC results against $l_{\infty}$-PGD attacks ($\epsilon=2/255$) on \emph{Bottle}, \emph{Grid}, \emph{Toothbrush}, \emph{Wood}  from MVTec AD at varying diffusion steps $k$ at inference time}
\label{table:gn}
\vskip -0.2in
\end{table}

\subsection{Impact of the Training Hyperparameters} \label{sec:training_hyper}
In this section, we conduct an ablation study to see how different training hyperparameters (i.e., noise schedule, diffusion timesteps for training, training iterations) will impact the standard/robust anomaly detection performance. Based on the results in Table \ref{table:exp_train_hyper}, we can observe that using the cosine noise schedule proposed by \citet{nichol2021improved} keeps a similar standard AUC while decreasing robust AUC by $3.7\%$ compared with using linear noise schedule and using larger $T$ also with more training iterations cannot significantly improve adversarial robustness and detection performance.

\begin{table}[htb]
\begin{center}
\resizebox{1.0\linewidth}{!}{
    \begin{tabular}{c c c c c }
    \toprule
        Iterations & $T$ & Noise schedule & Standard AUC (Avg.) & Robust AUC (Avg.)  \\
        \midrule
        $30K$ & $1000$ & linear & $96.7$ & $92.1$\\
        $30K$ & $1000$ & cosine & $96.3$ & $88.4$\\
        $30K$ & $2000$ & linear & $96.4$ & $91.3$\\
        $60K$ & $2000$ & linear & $96.6$ & $90.9$\\
    \bottomrule
    \end{tabular}}
\end{center}
\vskip -0.1in
\caption{Average standard AUC and robust AUC on \emph{Toothbrush}, \emph{Wood}, \emph{Hazelnut}, and \emph{Metal Nut} from MVTec AD, obtained by different training hyperparameters. }
\label{table:exp_train_hyper}
\vskip -0.2in
\end{table}

\subsection{More Results of the Comparison with Defense-enabled Anomaly Detectors} \label{sec:detail_defense_enabled}
\begin{table*}[tb]
    \begin{center}
    \resizebox{\linewidth}{!}{
        \begin{tabular}{c c c}
            \begin{tabular}{l | c c c c}
            \toprule
                Method & RAE & PLS & APAE & \textit{AdvRAD} \\
                \midrule
                Standard AUC & $36.1$ & $18.7$ & $36.6$ & $\mathbf{73.8}$ \\
                Robust AUC($l_{\infty}$-PGD) & $0$ & $3.0$ & $0$ & $\mathbf{70.5}$ \\
                Robust AUC($l_{2}$-PGD) & $29.9$ & $16.3$ & $30.6$ & $\mathbf{72.9}$ \\
            \bottomrule
            \end{tabular} &
            \begin{tabular}{l | c c c c}
            \toprule
                Method & RAE & PLS & APAE & \textit{AdvRAD} \\
                \midrule
                Standard AUC & $75.7$ & $7.9$ & $73.8$ & $\mathbf{100}$ \\
                Robust AUC($l_{\infty}$-PGD) & $61.3$ & $4.2$ & $59.5$ & $\mathbf{99.6}$ \\
                Robust AUC($l_{2}$-PGD) & $74.3$ & $7.7$ & $72.9$ & $\mathbf{99.9}$ \\
            \bottomrule
            \end{tabular} &
            \begin{tabular}{l | c c c c}
            \toprule
                Method & RAE & PLS & APAE & \textit{AdvRAD} \\
                \midrule
                Standard AUC & $61.0$ & $33.8$ & $42.0$ & $\mathbf{100}$ \\
                Robust AUC($l_{\infty}$-PGD) & $6.1$ & $10.1$ & $15.0$ & $\mathbf{97.8}$ \\
                Robust AUC($l_{2}$-PGD) & $54.3$ & $32.5$ & $39.9$ & $\mathbf{99.9}$ \\
            \bottomrule
            \end{tabular} \smallskip \\
            \smallskip 
            (a) \emph{Carpet} & (b) \emph{Grid} & (c) \emph{Leather}\\
            \begin{tabular}{l | c c c c}
            \toprule
                Method & RAE & PLS & APAE & \textit{AdvRAD} \\
                \midrule
                Standard AUC & $70.5$ & $35.4$ & $71.7$ & $\mathbf{95.4}$ \\
                Robust AUC($l_{\infty}$-PGD) & $23.5$ & $22.5$ & $26.0$ & $\mathbf{93.9}$ \\
                Robust AUC($l_{2}$-PGD) & $67.4$ & $34.5$ & $68.9$ & $\mathbf{94.8}$ \\
            \bottomrule
            \end{tabular} &
            \begin{tabular}{l | c c c c}
            \toprule
                Method & RAE & PLS & APAE & \textit{AdvRAD} \\
                \midrule
                Standard AUC & $80.9$ & $80.4$ & $94.3$ & $\mathbf{98.3}$ \\
                Robust AUC($l_{\infty}$-PGD) & $20.5$ & $39.8$ & $56.6$ & $\mathbf{95.2}$ \\
                Robust AUC($l_{2}$-PGD) & $76.8$ & $79.1$ & $92.2$ & $\mathbf{95.5}$ \\
            \bottomrule
            \end{tabular} &
            \begin{tabular}{l | c c c c}
            \toprule
                Method & RAE & PLS & APAE & \textit{AdvRAD} \\
                \midrule
                Standard AUC & $70.2$ & $60.6$ & $90.0$ & $\mathbf{99.6}$ \\
                Robust AUC($l_{\infty}$-PGD) & $0$ & $8.7$ & $55$ & $\mathbf{96.9}$ \\
                Robust AUC($l_{2}$-PGD) & $35.6$ & $45.2$ & $87.8$ & $\mathbf{97.5}$ \\
            \bottomrule
            \end{tabular} \smallskip \\
            \smallskip 
            (a) \emph{Tile} & (b) \emph{Wood} & (c) \emph{Bottle}\\
            \begin{tabular}{l | c c c c}
            \toprule
                Method & RAE & PLS & APAE & \textit{AdvRAD} \\
                \midrule
                Standard AUC & $64.9$ & $53.6$ & $67.4$ & $\mathbf{79.2}$ \\
                Robust AUC($l_{\infty}$-PGD) & $\mathbf{44.3}$ & $18.0$ & $42.2$ & $38.9$ \\
                Robust AUC($l_{2}$-PGD) & $63.1$ & $47.8$ & $65.5$ & $\mathbf{65.7}$ \\
            \bottomrule
            \end{tabular} &
            \begin{tabular}{l | c c c c}
            \toprule
                Method & RAE & PLS & APAE & \textit{AdvRAD} \\
                \midrule
                Standard AUC & $50.9$ & $34.9$ & $63.1$ & $\mathbf{90.5}$ \\
                Robust AUC($l_{\infty}$-PGD) & $16.7$ & $6.9$ & $9.1$ & $\mathbf{53.5}$ \\
                Robust AUC($l_{2}$-PGD) & $46.2$ & $30.9$ & $56.9$ & $\mathbf{68.1}$ \\
            \bottomrule
            \end{tabular} &
            \begin{tabular}{l | c c c c}
            \toprule
                Method & RAE & PLS & APAE & \textit{AdvRAD} \\
                \midrule
                Standard AUC & $31.8$ & $64.3$ & $63.8$ & $\mathbf{97.3}$ \\
                Robust AUC($l_{\infty}$-PGD) & $0$ & $29.4$ & $31.6$ & $\mathbf{91.5}$ \\
                Robust AUC($l_{2}$-PGD) & $11.4$ & $59.6$ & $60$ & $\mathbf{94.3}$ \\
            \bottomrule
            \end{tabular} \smallskip \\
            \smallskip 
            (a) \emph{Cable} & (b) \emph{Capsule} & (c) \emph{Hazelnut}\\
            \begin{tabular}{l | c c c c}
            \toprule
                Method & RAE & PLS & APAE & \textit{AdvRAD} \\
                \midrule
                Standard AUC & $36.5$ & $66.9$ & $41$ & $\mathbf{91.0}$ \\
                Robust AUC($l_{\infty}$-PGD) & $0$ & $21.4$ & $22.8$ & $\mathbf{85.9}$ \\
                Robust AUC($l_{2}$-PGD) & $29.4$ & $61.7$ & $39.1$ & $\mathbf{87.9}$ \\
            \bottomrule
            \end{tabular} &
            \begin{tabular}{l | c c c c}
            \toprule
                Method & RAE & PLS & APAE & \textit{AdvRAD} \\
                \midrule
                Standard AUC & $67.5$ & $41.1$ & $66.6$ & $\mathbf{94.4}$ \\
                Robust AUC($l_{\infty}$-PGD) & $10$ & $3.9$ & $9.3$ & $\mathbf{39}$ \\
                Robust AUC($l_{2}$-PGD) & $59.5$ & $13.0$ & $59.7$ & $\mathbf{80.3}$ \\
            \bottomrule
            \end{tabular} &
            \begin{tabular}{l | c c c c}
            \toprule
                Method & RAE & PLS & APAE & \textit{AdvRAD} \\
                \midrule
                Standard AUC & $0.1$ & $0.8$ & $45.7$ & $\mathbf{99.3}$ \\
                Robust AUC($l_{\infty}$-PGD) & $0$ & $0.8$ & $12.5$ & $\mathbf{87.6}$ \\
                Robust AUC($l_{2}$-PGD) & $0$ & $0.8$ & $41.2$ & $\mathbf{91.8}$ \\
            \bottomrule
            \end{tabular} \smallskip \\
            \smallskip 
            (a) \emph{Metal Nut} & (b) \emph{Pill} & (c) \emph{Screw}\\
            \begin{tabular}{l | c c c c}
            \toprule
                Method & RAE & PLS & APAE & \textit{AdvRAD} \\
                \midrule
                Standard AUC & $81.4$ & $72.8$ & $81.7$ & $\mathbf{100}$ \\
                Robust AUC($l_{\infty}$-PGD) & $26.4$ & $36.7$ & $27.2$ & $\mathbf{95.8}$ \\
                Robust AUC($l_{2}$-PGD) & $76.7$ & $70$ & $76.4$ & $\mathbf{99.4}$ \\

            \bottomrule
            \end{tabular} &
            \begin{tabular}{l | c c c c}
            \toprule
                Method & RAE & PLS & APAE & \textit{AdvRAD} \\
                \midrule
                Standard AUC & $76.8$ & $67.7$ & $72.3$ & $\mathbf{90.6}$ \\
                Robust AUC($l_{\infty}$-PGD) & $38.8$ & $27.7$ & $55.8$ & $\mathbf{74.5}$ \\
                Robust AUC($l_{2}$-PGD) & $73.3$ & $62.9$ & $70.7$ & $\mathbf{84.3}$ \\
            \bottomrule
            \end{tabular} &
            \begin{tabular}{l | c c c c}
            \toprule
                Method & RAE & PLS & APAE & \textit{AdvRAD} \\
                \midrule
                Standard AUC & $52.8$ & $57.3$ & $60.6$ & $\mathbf{99.7}$ \\
                Robust AUC($l_{\infty}$-PGD) & $4.0$ & $6.9$ & $26.2$ & $\mathbf{96.2}$ \\
                Robust AUC($l_{2}$-PGD) & $48.5$ & $50.4$ & $56.8$ & $\mathbf{99.2}$ \\
            \bottomrule
            \end{tabular} \smallskip \\
            \smallskip 
            (a) \emph{Toothbrush} & (b) \emph{Transistor} & (c) \emph{Zipper}\\
        \end{tabular}}
    \end{center}
\vskip -0.1in
\caption{Standard AUC and robust AUC on 15 sub-datasets of MVTec AD against $l_{\infty}$-PGD ($\epsilon=2/255$), $l_{2}$-PGD ($\epsilon=0.2$) attacks, obtained by RAE \citep{zhou2017anomaly}, PLS\citep{lo2022adversarially}, APAE \citep{goodge2020robustness} and ours.}
\label{table:details_comp_defense_enabled}
\vskip -0.1in
\end{table*}
We report the comparison results of our method and defense-enabled anomaly detection methods against PGD attacks on 15 categories of MVTec AD benchmark in Table \ref{table:details_comp_defense_enabled}. We can observe that our method obtains the best robust AUC scores on 14 of 15 categories against $l_{\infty}$-PGD ($\epsilon=2/255$) attacks and outperforms all baselines regarding the standard AUC on clean data and robust AUC against $l_{2}$-PGD ($\epsilon=0.2$) attacks on all sub-datasets.

\end{document}